\documentclass[10pt,twocolumn,letterpaper]{article}

\usepackage{cvpr}              

\newcommand{\revision}[1]{{\color{black}#1}}
\newcommand{\model}{Qwen-3D}

\usepackage{multirow}

\usepackage{xcolor} 
\usepackage{colortbl}     
\usepackage{pifont}       
\usepackage[T1]{fontenc}

\definecolor{LightGray}{gray}{0.95}

\definecolor{cvprblue}{rgb}{0.21,0.49,0.74}
\usepackage[pagebackref,breaklinks,colorlinks,allcolors=cvprblue]{hyperref}

\def\paperID{00027} 
\def\confName{CVPR}
\def\confYear{2026}

\title{Qwen-3D: A Generalist 3D Vision-Language Model for Spatial Understanding}

\author{
		Lucy Lin$^{\dagger}$, Ayush Jain$^{\dagger}$, Yifan Liu, Katerina Fragkiadaki \\
		Carnegie Mellon University
		\\
		{\tt \small{\{lucylin,ayushj2,yifanliu,kfragki2\}@andrew.cmu.edu}}\\
	}

\newcommand\blfootnote[1]{%
  \begingroup
  \renewcommand\thefootnote{}\footnote{#1}%
  \addtocounter{footnote}{-1}%
  \endgroup
}

\begin{document}
\maketitle
\blfootnote{$^{\dagger}$Equal contribution}

\begin{abstract}
Large Multimodal Models (LMMs) have achieved remarkable success on images and short videos, yet scaling them to long videos remains challenging due to frame-centric tokenization and limited context windows. 3D geometry provides a natural compression mechanism for visual streams: depth and camera pose enable observations from multiple views and time steps to be fused into a persistent, world-aligned representation.  
While recent 3D LMMs leverage geometry-aware representations to improve spatial reasoning, they continue to lag behind specialist 3D perception systems on grounding and segmentation tasks. We argue that a key limitation is geometry-aware decoding: existing methods communicate 3D predictions through language tokens, proposal selection, or lightweight grounding queries, creating a bottleneck between language reasoning and dense geometric prediction. 
Building on these insights, we introduce \model{}, a geometry-aware LMM that compresses visual information within the Qwen backbone using multi-view geometric cues, enabling efficient long-horizon visual reasoning over static scenes. \model{} augments visual tokens with 3D Rotary Positional Embeddings, allowing attention to operate directly in 3D scene space rather than across independent image frames and thereby facilitating scalable cross-view and temporal reasoning. To bridge language and geometry, \model{} incorporates a query-based segmentation decoder that grounds language directly in the underlying 3D scene representation, unifying referential grounding, instance segmentation, and visual question answering across both images and videos. Across a diverse set of benchmarks, \model{} surpasses existing 3D LMMs and outperforms several large proprietary 2D models. Notably, \model{} achieves these improvements while maintaining strong performance on standard 2D vision--language benchmarks by jointly training on 2D and 3D data. Our code and checkpoints can be found at the project website \url{https://qwen-3d.github.io/}.
\end{abstract}
    
\section{Introduction}
\label{sec:intro}
\begin{figure*}[!ht]
\centering
    \includegraphics[width=\textwidth]{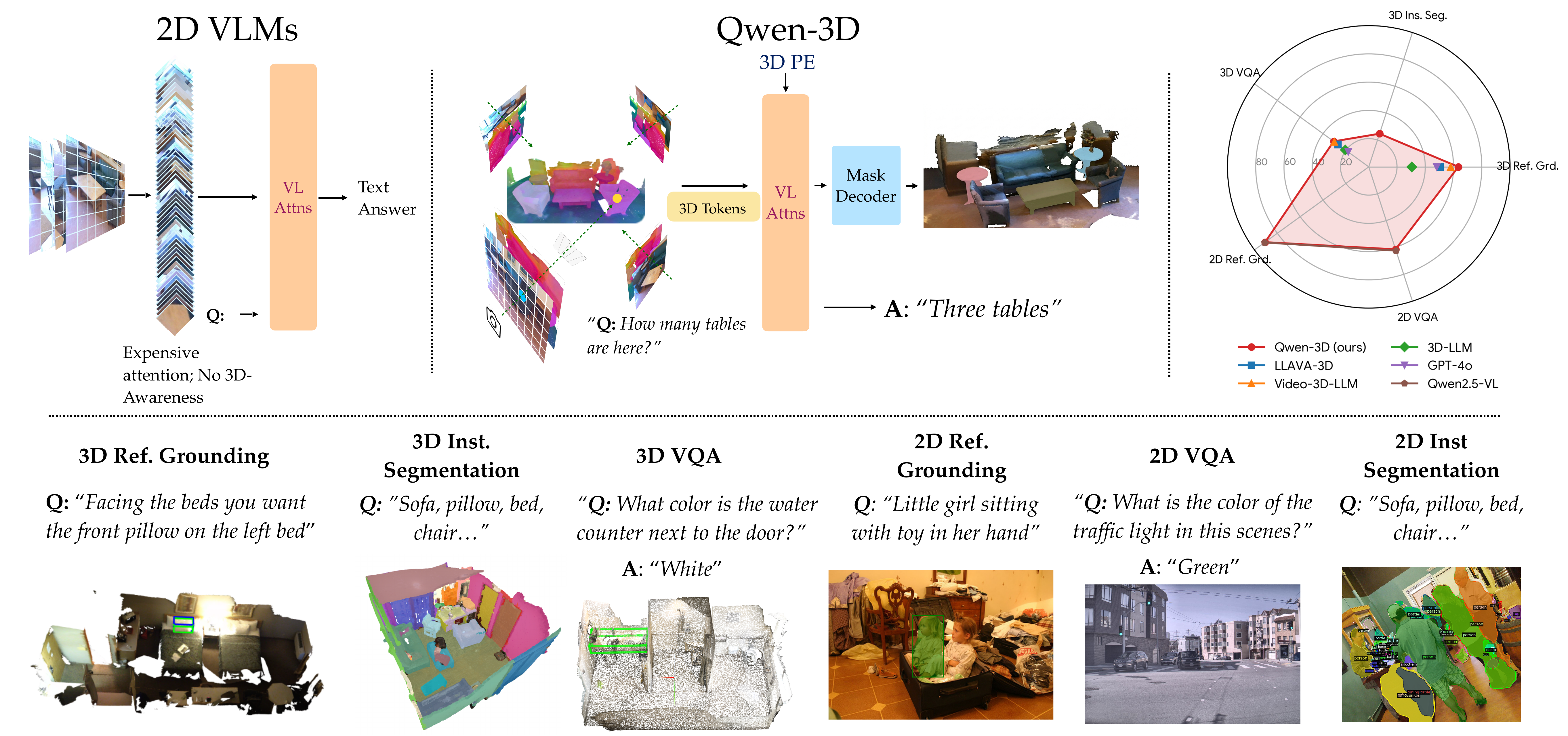}
    \caption{\textbf{Qwen-3D performs attention directly in 3D world space rather than over independent image frames.} Given multi-view RGB observations, depth, and camera poses, Qwen-3D maps visual tokens into a shared 3D coordinate system and applies geometry-aware attention through 3D Rotary Positional Embeddings. The model jointly supports language reasoning, 2D grounding, and 3D grounding within a unified architecture, achieving state-of-the-art performance across a broad range of vision–language and 3D understanding benchmarks. \vspace{-2em}}
    \label{fig:teaser}
\end{figure*}

Current Vision–Language Models (VLMs) perform well on images and short video clips, but struggle with long multi-view video streams. Processing long sequences is computationally expensive due to the quadratic cost of attention, and limited context windows prevent long-range spatio-temporal reasoning across frames. Multi-view 3D geometry, in the form of depth and camera poses, offers a  principled alternative, allowing video frames to be mapped into a shared 3D coordinate system. This enables compression of long multi-view streams into compressed persistent scene representations where temporally distant frames may correspond to nearby 3D locations. 

Existing approaches to integrating such 3D information compression into VLMs in order to improve their long range reasoning abilities generally follow one of two distinct paradigms. One line of work introduces 3D point clouds as auxiliary inputs to the model~\cite{leo,chatscene,ll3da}, either alongside or in place of multi-view images. While these approaches expose explicit geometric structure, they treat point clouds as a modality separate from the visual tokens,  overlooking the fact that point cloud features are inherently aligned with image features with corresponding depth. The second line of work integrates geometry directly into the visual token representation. Methods such as LLaVA-3D\cite{l3d} and Video-3D-LLM\cite{video3dllm} modify positional encodings such that multi-view image tokens are embedded according to their 3D world coordinates rather than their 2D image-plane positions. This approach allows the model to reason over multi-view observations in a shared spatial coordinate system while maintaining the strong visual representations learned by large VLM backbones.

Despite these advances, existing 3D large multimodal models (LMMs) still lag substantially behind specialist 3D perception systems. Dedicated models trained for detection, segmentation, and grounding continue to outperform general-purpose 3D LMMs by a large margin~\cite{3dvista,odin,univlg}. Moreover, most current 3D LMMs do not even attempt standard 3D perception tasks such as object detection on ScanNet~\cite{scannet200,scannet}. The only exception, Grounded-3D-LLM~\cite{grounded3dllm}, achieves less than half the performance of state-of-the-art 3D detectors.

We argue that this gap stems from a fundamental challenge in adapting language-centric architectures to 3D perception. While large language models excel at reasoning over discrete tokens, dense 3D grounding requires predicting spatially precise outputs in a continuous world coordinate system. Unlike images, which provide a canonical pixel coordinate frame, 3D scenes admit no universal reference frame: the same object may appear at entirely different coordinates across scans and environments. As a result, autoregressively decoding 3D boxes, coordinates, or masks as language tokens is an unnatural interface for 3D perception.

\begin{figure*}[t!]
\centering
    \includegraphics[width=\textwidth]{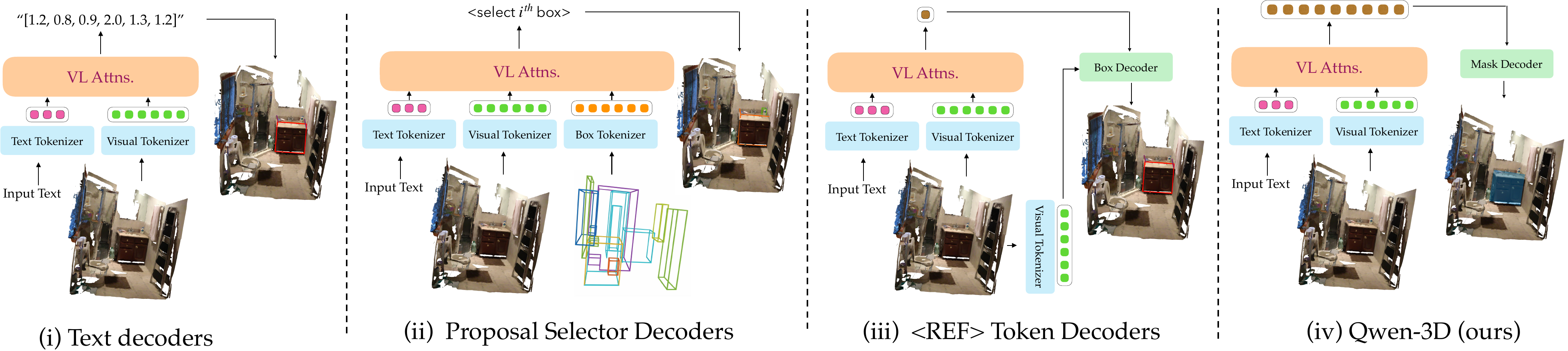}
    \caption{\textbf{Decoding object information in 3D LMMs.} From left to right: (i) \emph{Text-space decoding}, as in 3D-LLM~\cite{3dllm}, where object coordinates or bounding boxes are generated autoregressively as language tokens. Unlike images, 3D scenes do not admit a canonical world coordinate frame, making dense spatial prediction through text generation inherently ambiguous. (ii) \emph{Proposal-selection decoding}, as in Video-3D-LLM~\cite{video3dllm,chatscene,leo}, where the model selects from a predefined set of object proposals. This approach is fundamentally limited by proposal quality and cannot localize objects absent from the proposal set. (iii) \emph{Special-token decoding}, as in LLaVA-3D~\cite{l3d,grounded3dllm}, where grounding information is communicated through a single dedicated token. While effective for coarse language-to-vision communication, this creates a severe information bottleneck between language reasoning and dense geometric prediction. (iv) \textbf{\model{} (ours)}, which directly connects contextualized vision–language features from the backbone to a mask-based segmentation decoder. By sharing dense visual representations between language reasoning and geometric prediction, \model{} grounds multiple referential expressions directly in the underlying scene representation, enabling high-quality 2D and 3D grounding and segmentation.}
    \label{fig:decoders}
\end{figure*}

Existing 3D LMMs typically address this problem either by representing grounding outputs through text generation or by attaching lightweight grounding modules that communicate with the backbone through a small set of query vectors (Figure \ref{fig:decoders}). While these approaches preserve the reasoning capabilities of the underlying language model, they create a severe information bottleneck between high-capacity visual representations and the dense geometric predictions required for grounding. Consequently, current 3D LMMs improve high-level spatial reasoning but remain significantly weaker than specialist systems on core 3D perception tasks. This observation raises an important open question: how should a large multimodal model interface with a 3D grounding system?

We introduce \model{}, a geometry-aware 3D LMM that extends the Qwen family of models~\cite{qwen2.5} with explicit mechanisms for multi-view reasoning and object grounding (Figure \ref{fig:teaser}). Rather than treating grounding as language generation or restricting communication through a small set of latent queries, we propose a unified architecture in which the language model and grounding decoder share dense visual representations. This design enables rich bidirectional interactions between language reasoning and geometric perception, substantially improving grounding accuracy while preserving the general-purpose capabilities of the multimodal backbone. 
Built on the strong 2D foundation of Qwen2.5-VL, \model{} integrates 3D structure directly into the vision–language backbone. First, we leverage geometric information to compress visual tokens within the backbone, merging tokens originating from nearby 3D locations. This enables efficient reasoning over long multi-view videos while preserving spatial consistency across views. Second, we introduce 3D Rotary Positional Embeddings to the Qwen backbone, allowing attention within the backbone to operate in a geometry-aware coordinate system and improving cross-view spatial reasoning.

Across a wide range of 3D grounding benchmarks, \model{} outperforms both proprietary 2D VLMs and the strongest existing 3D LMMs while maintaining strong performance on 2D tasks. Compared to the previous 3D LMM state-of-the-art, \model{} improves 3D visual grounding by 4\% Acc@25, surpasses the best single-stage 3D LMMs by 12\% Acc@25, and increases 3D instance segmentation accuracy by 13\% mAP. The model also achieves competitive performance on 3D visual question answering while preserving the strong 2D capabilities of its backbone. Furthermore, \model{} substantially narrows the gap between general-purpose 3D LMMs and specialist 3D grounding models on in-domain benchmarks, while significantly outperforming specialist methods on out-of-distribution 3D scenes and language instructions. Following prior 3D LMMs such as LLaVA-3D~\cite{l3d} and Grounded-3D-LLM~\cite{grounded3dllm}, we characterize \model{} as a vision–language generalist based on its broad multi-modal 2D and 3D capabilities, while further broadening the evaluation scope beyond these baselines to include 3D object detection, text-only baselines, and generalization across unseen distributions.

\noindent\textbf{Contributions.} Our contributions are as follows:
\begin{itemize}
    \item A geometry-aware VLM backbone that integrates multi-view structure with 3D rotary positional embeddings and geometry-based token compression.
    
    \item A unified 3D grounding architecture with a query-based segmentation decoder that grounds language directly in world space and shares full visual token representations with the VLM backbone.
    
    \item A general-purpose 3D multimodal training framework that jointly learns from 2D and 3D data, preserving strong vision–language capabilities while improving 3D spatial reasoning and grounding.
    
    \item State-of-the-art performance among 3D-LMM methods across 3D benchmarks, improving 3D visual grounding by 4\% and instance segmentation by 13\%, while maintaining strong 2D multimodal performance.
\end{itemize}

We make our code publicly available at \url{https://qwen-3d.github.io/}.

\section{Related Work}

\label{sec:related}
Building upon the rapid progress of 2D Large Multimodal Models~\cite{qwen2.5,gemini,openai2024gpt4o}, recent work has focused on extending these models to understand 3D scenes. Existing approaches can be grouped into four main categories:

\noindent \textbf{(a) 3D point cloud encoders trained from scratch.} 
Methods such as LL3DA~\cite{ll3da}, Scene-LLM~\cite{scenellm}, and Grounded-3D-LLM~\cite{grounded3dllm} augment multi-view image streams with explicit 3D point cloud encoders. The resulting 3D features are projected into an LLM backbone in addition to or as a replacement for 2D image features. These approaches, however, require large-scale point cloud–language datasets for alignment—an acute limitation given the scarcity of 3D data. In contrast, \model{} builds upon powerful 2D pre-trained features and augments them with 3D information via positional encodings.

\noindent \textbf{(b) Learnable 3D feature compression.} 
3D-LLM~\cite{3dllm} uses Q-Former layers~\cite{blip2} to compress large numbers of 2D foundation-model features into small sets of latent tokens. In contrast, \model{} performs multi-view feature compression in a parameter-free manner, directly guided by the 3D spatial layout of the tokens. 

\noindent \textbf{(c) Object-centric approaches.} Another line of work uses object-level features as input to the language model. Approaches such as LEO~\cite{leo} and ChatScene~\cite{chatscene} first detect objects with off-the-shelf 2D or 3D detectors, pool features within detected regions, then feed these pooled features into their VLMs. These methods may yield structured object representations and improve grounding, but performance is fundamentally constrained by the robustness of the detectors themselves, which often struggle due to limited data diversity. In contrast, \model{} is a single-stage model that directly grounds the language in the 3D visual stream.

\noindent \textbf{(d) Positional embedding adaptation.} Several methods modify the positional embeddings of multi-view visual tokens to better encode 3D spatial relationships~\cite{odin,univlg,l3d,video3dllm}. Our model follows this general paradigm. Similar to these models, we incorporate 3D information via positional embeddings in the vision-language attention. We utilize 3D Rotary Positional Encoding for this purpose. 

Beyond spatial encoding, \model{} also differs from prior work in how grounded outputs are decoded. We discuss these grounding architectures in detail in the following section.





\vspace{-1em}
\paragraph{3D Visual Grounding.}
Visual grounding—identifying objects referred to by language—is a fundamental capability for 3D vision–language systems. Early work~\cite{butd,locate3d,sps3d} achieved strong performance by designing specialized architectures tailored for 3D grounding. Subsequent methods~\cite{3dvista,pq3d,univlg,locate3d} unified grounding, question answering, and captioning within a single framework. More recently, 3D LMMs have leveraged large-scale pretrained vision–language features to assist grounding in 3D scenes; our method follows this paradigm. These newer approaches typically adopt one of three designs (~\cref{fig:decoders}):

\noindent \textbf{(a) Direct bounding-box decoding. (\cref{fig:decoders}a)} 
Models such as 3D-LLM~\cite{3dllm} directly decode 3D bounding boxes in the text space. However, they achieve low performance on localization tasks, likely due to the scarcity of 3D-language data and the unstructured nature of 3D scenes. 

\noindent\textbf{(b) Two-stage grounding via proposal selection. (\cref{fig:decoders}b)} Approaches such as Video-3D-LLM~\cite{video3dllm}, ChatScene~\cite{chatscene}, and LEO~\cite{leo} first run an object detector to generate candidate proposals, then select the object that best matches the query. While this improves grounding robustness, performance is bottle-necked by the quality of the proposals.

\noindent\textbf{(c) Special token decoding. (\cref{fig:decoders}c)} 
Following the mask-as-embedding paradigm introduced by LISA~\cite{lisa}, several methods decode a special grounding token (e.g. $<\mathrm{REF}>$) and localize it using an explicit decoder head, as in LLaVA-3D~\cite{l3d}, Grounded-3D-LLM~\cite{grounded3dllm}, and Reason3D~\cite{reason3d}. Essentially, the VLM heads and the decoder heads are only connected via the generated $<\mathrm{REF}>$ tokens. Although these methods can, in principle, ground multiple instances from a category, only Grounded-3D-LLM has been applied to 3D object detection.

Rather than relying on text-space bounding boxes, proposal selection, or grounding tokens, \textbf{\model{} directly connects contextualized vision–language features from the VLM backbone to a mask-based segmentation decoder} (~\cref{fig:decoders}d). This avoids the bottleneck imposed by the special token decoding and aligns more naturally with contemporary multi-object detection and segmentation architectures. As a result, \model{} achieves state-of-the-art grounding and detection performance among 3D LMMs.

\section{Method}
\label{sec:method}

\begin{figure*}[h!]
\centering
    \includegraphics[width=0.9\textwidth]{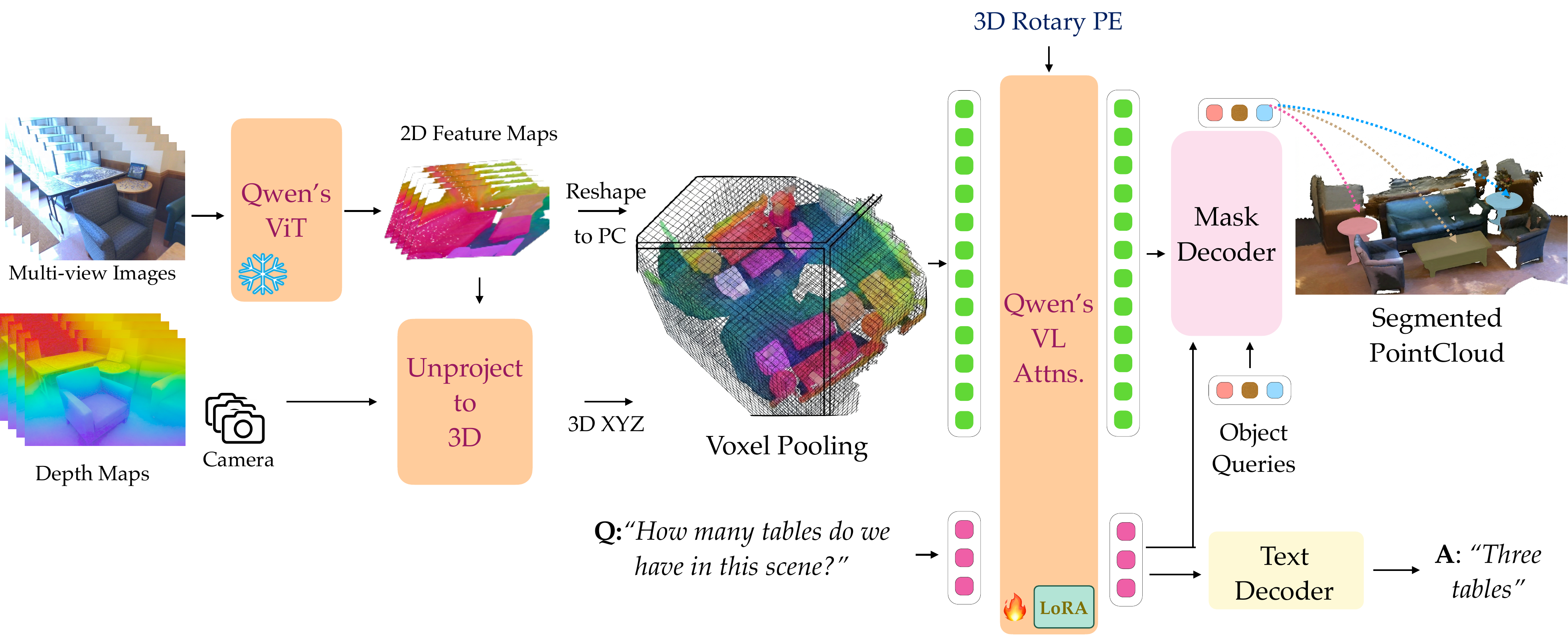}
    \caption{\textbf{\model{} architecture.} Given a natural language query and multi-view RGB-D  inputs, the Qwen2.5-VL vision encoder extracts multi-view 2D features, unprojects them into world-space XYZ coordinates, and voxel-pools them to reduce spatial redundancy. The resulting tokens are fused with text via Qwen vision–language attention layers augmented with 3D Rotary Positional Embeddings and LoRA adaptation. Two task heads operate on the shared tokens: a 3D mask decoder for referential grounding and instance segmentation and a text decoder for open-ended question answering.}
    \label{fig:method}
\end{figure*}

Qwen-3D extends Qwen2.5-VL with three key components. First, we construct a geometry-aware scene representation by projecting multi-view visual features into a shared 3D coordinate system and compressing redundant observations through voxel-based token merging. Second, we enable geometry-aware reasoning by replacing image-plane positional encodings with 3D Rotary Positional Embeddings, allowing attention to operate directly in world space. Third, we introduce a geometry-aware decoding mechanism that directly couples contextualized vision--language features with a mask-based grounding decoder, enabling dense language-guided prediction in both 2D and 3D. Figure~\ref{fig:method} provides an overview.

\subsection{Geometry-Aware Scene Representation}

Given a set of posed RGB-D observations, we first lift visual features into a shared world coordinate system. Specifically, we use the Qwen2.5-VL vision encoder to extract image features and unproject their corresponding depth values using camera intrinsics and poses, producing a set of aligned feature-coordinate pairs

\[
\mathcal{FC} = \{(f_i,p_i)\}_{i=1}^{M},
\]
where \(f_i \in \mathbb{R}^{D}\) denotes a visual feature and \(p_i \in \mathbb{R}^{3}\) its corresponding world-space coordinate.



Multi-view observations produce substantial redundancy because pixels from different views often correspond to the same physical location. Rather than reasoning over frame-level tokens, we aggregate observations directly in world space. 
Specifically, we apply voxel-based token merging (with a voxel size of 5cm) to the feature-coordinate pairs, following~\cite{odin,univlg,l3d}. This process discretizes the space and mean-pools the features and coordinates within each occupied voxel, yielding a compact, unordered set of geometry-aligned tokens $\mathcal{FC}' = \{(f_j, p_j)\}_{j=1}^{M'}$, where $M' \leq M$.

\subsection{Geometry-Aware Vision--Language Attention}
\label{para:vl_atten}
Unlike conventional VLMs that perform attention over image-plane coordinates, Qwen-3D performs attention directly in a shared world coordinate system. 
We use the Qwen2.5-VL language tokenizer~\cite{qwen2.5} to embed the input natural language query into a sequence of tokens 
$\mathcal{T} = \{t_k\}_{k=1}^{L}$, 
where $t_k \in \mathbb{R}^{D}$ and $L$ is the number of tokens. 
The concatenated sequence of voxelized 3D point features and text tokens is then processed by $N$ pre-trained multimodal attention layers from Qwen2.5-VL. To enable spatial reasoning, we adapt Qwen’s Multimodal RoPE—originally designed for 1D or 2D grids—to encode 3D world coordinates $(x, y, z)$, resulting in four positional components 
\vspace{-1em}

\[
\text{PE}_{3D}(\mathbf{p}) = [\text{PE}(t); \text{PE}(x); \text{PE}(y); \text{PE}(z)].
\]

where $t$ represents a token’s temporal position in the sequence and $x, y, z$ denote 3D spatial coordinates. For text, the temporal ID increments sequentially (reducing to standard 1D RoPE), while point cloud tokens share a constant temporal ID and use their world coordinates as spatial IDs.

This embedding defines a rotation matrix $R(\mathbf{p})$ applied to query and key vectors before attention:
$\tilde{\mathbf{q}} = R(\mathbf{p}_q)\mathbf{q}, \;
\tilde{\mathbf{k}} = R(\mathbf{p}_k)\mathbf{k}$

Because 3D point clouds are permutation-invariant, we replace Qwen2.5-VL's autoregressive causal masking with full attention over visual tokens. This improves 3D performance without degrading the backbone's original capabilities. Further attention masking details and RoPE ablations are provided in the appendix and experiments section, respectively.

\subsection{Geometry-Aware Decoding}

Most existing 3D LMMs communicate grounding information through language tokens, proposal indices, or a small number of dedicated grounding tokens~\cite{3dllm,video3dllm,l3d,grounded3dllm}. While these interfaces preserve the reasoning capabilities of the language model, they create a bottleneck between the rich visual representations learned by the backbone and the dense geometric predictions required for grounding. Instead, we directly expose the full contextualized vision--language representation produced by the backbone to a mask-based segmentation decoder.

Let
\[
\mathcal{V} = \{v_i, i=1..M'\}
\]
denote the contextualized visual tokens produced by the geometry-aware attention layers and
\[
\mathcal{T} = \{t_j, j=1..L\}
\]
the corresponding language tokens. We instantiate (N) learnable object queries
\[
\mathcal{Q} = \{q_n ,n=1..N\}
\]
and process them through a Mask2Former-style decoder~\cite{m2f}. Each decoder layer alternates between cross-attention to visual and language tokens, and self-attention among object queries. This allows each query to jointly reason about scene geometry and linguistic context while maintaining access to the full visual representation rather than a compressed grounding bottleneck.

For all attention involving visual tokens, we use positional embeddings corresponding to the underlying modality (2D image coordinates or 3D world coordinates). After the final decoder layer, each object query predicts (i) a segmentation mask through a dot product with the updated visual tokens and (ii) a language grounding score through a dot product with the language tokens. 
This unified decoder naturally supports both 2D and 3D grounding. For referential grounding, object queries learn to associate text spans with the corresponding visual regions. For question answering and captioning, we use the original Qwen2.5-VL language head to perform next-token prediction over the contextualized scene--language representation. 
This design enables dense geometric prediction while preserving the reasoning capabilities of the underlying language model.

\vspace{-1em}
\paragraph{Training Objectives} 

We supervise \model{} on three losses: 
\noindent\textbf{(a) Mask loss:} We assign predictions to ground-truth instances via Hungarian matching~\cite{Carion2020EndtoEndOD} and supervise matched masks with Binary Cross-Entropy (BCE) and Dice loss, following Mask2Former~\cite{m2f}.

\noindent\textbf{(b) Text-span grounding loss:} As in~\cite{glip, Kamath2021MDETRM, butd}, we supervise each predicted text span with the corresponding matched ground-truth text span using Binary Cross Entropy loss. The queries that remain unmatched are supervised to predict low probability over all text tokens. 

\noindent\textbf{(c) Text generation loss:} For question-answering and captioning tasks, we use a token-level cross-entropy on the generated answer.

\noindent Our complete loss is formulated as:
\vspace{-1em}

\begin{align}
    \mathcal{L} &= \alpha_{\text{mask}}\,\mathcal{L}_{\text{mask}} 
                + \alpha_{\text{textground}}\,\mathcal{L}_{\text{textground}}
                + \alpha_{\text{gen}}\,\mathcal{L}_{\text{gen}}
\end{align}

where $\mathcal{L}_{\text{mask}}$ is the mask loss, $\mathcal{L}_{\text{textground}}$ is the text grounding loss, $\mathcal{L}_{\text{gen}}$ is the text generation loss, and the $\alpha$'s are the loss weight terms.

\subsection{Joint 2D-3D Training}
\model{} shares parameters across 2D and 3D modalities, natively processing single or multiple RGB images as well as multi-view posed RGB-D frames. Within both the VLM backbone and the grounding decoder, 2D visual tokens are structured on a regular grid with 2D positional embeddings, whereas 3D inputs are represented as unordered point tokens with 3D embeddings. To improve alignment between 2D and 3D representations during training, we lift 2D data to 3D with probability $p$ using reconstruction models such as MoGE~\cite{moge} following UniVLG~\cite{univlg}. This enables a single set of parameters to operate seamlessly across images, videos, and reconstructed 3D scenes.

\vspace{-1em}
\paragraph{Implementation Details}
\model{} introduces only $\sim$50M trainable parameters. We freeze the Qwen2.5-VL ViT backbone, fine-tune the vision--language attention layers via LoRA~\cite{lora}, and train the mask decoder from scratch. We train jointly on 2D and 3D datasets for 200k iterations (learning rate $10^{-4}$) on eight 48GB L40S GPUs with an effective batch size of 8, which takes approximately three days. Text-generation loss is applied exclusively to captioning and question-answering tasks; for detection, we construct prompts~\cite{butd} by concatenating object class names (e.g., ``find chair. table. sofa.''). 

Following~\cite{univlg,odin}, we subsample 15 frames per scene during training. At inference time, we feed all posed RGB-D frames to our model ($\sim$90 on average for ScanNet), which takes about $2$ seconds per scene end-to-end. Our voxel-pooling strategy is critical for enabling this scalability - without it, the model runs out of memory even with substantially fewer input frames. We include additional hyperparameters and an ablation of the test-time frame subsampling in the appendix.

\section{Experiments}

\begin{table*}[!ht]
    \centering
    \vspace{-0.5em}
    \caption{\textbf{Results on 3D Visual Grounding and VQA for both experts and LMMs.} $^*$2D VLM numbers obtained from prior works ~\cite{locate3d,zerogpt,viewongraph,vgllm}}
    \vspace{2pt}
    \resizebox{.8\textwidth}{!}{
    \begin{tabular}{ll*{7}{c}}
        \toprule
        & & \multicolumn{2}{c}{\textbf{ScanRefer}} 
        & \multicolumn{4}{c}{\textbf{ScanQA (Val)}} 
        & \multicolumn{1}{c}{\textbf{SQA3D (Test)}}\\

        & \textbf{Methods} 
        & \begin{tabular}{@{}c@{}}Acc \\ @25\\ \end{tabular} 
        & \begin{tabular}{@{}c@{}}Acc \\ @50\\ \end{tabular} 

        & \begin{tabular}{@{}c@{}}EM@1\end{tabular}
        & \begin{tabular}{@{}c@{}}C\end{tabular} 
        & \begin{tabular}{@{}c@{}}M\end{tabular} 
        & \begin{tabular}{@{}c@{}}R\end{tabular} 

        & \begin{tabular}{@{}c@{}}EM@1\end{tabular} \\
        \midrule

        & BUTD-DETR~\cite{butd} & 52.2 & 39.8  & - & - & - & - & - \\
        & ScanQA~\cite{scanqa} & - & -  & 23.5 & 67.3 & 13.6 & 34.3 & - \\
        & PQ3D~\cite{pq3d} & 56.7 & 51.8  & 20.0 & 65.2 & 13.9 & - & 47.1 \\
        \textbf{Experts}
        & 3D-VisTA~\cite{3dvista} & 51.0 & 46.2  & 22.4 & 69.6 & 13.9 & 35.7 & 48.5 \\
        & ODIN~\cite{odin} & 43.1 & 33.4  & - & - & - & - & - \\
        & Locate-3D~\cite{locate3d} & 61.1 & 50.9 & - & - & - & - & - \\
        & UniVLG~\cite{univlg} & 63.5 & 56.4 & 25.7 & 78.5 & 15.2 & 40.0 & 50.2 \\

        \midrule

        &\textbf{2D VLMs}$^*$ & & & & & & & \\
        & LLaMA~\cite{llama}  & 37.3 & 24.2 & - & - & - & - & - \\
        & Qwen2-VL-2B~\cite{qwen2vl} & 35.9 & 31.9 & - & - & - & - & - \\
        & Qwen2-VL-72B~\cite{qwen2vl}  & 44.8 & 40.3 & - & - & - & - & - \\
        & Qwen2.5-VL-3B~\cite{qwen2.5llm}  & 36.4 & 11.8 & - & - & - & - & - \\
        & Qwen2.5-VL-7B~\cite{qwen2.5llm}  & 41.6 & 14.9 & - & - & - & - & - \\
        & GPT-4o~\cite{openai2024gpt4o}  & 48.2 & 32.5 & 18.0 & 58.3 & 14.2 & 33.4 & - \\
        
        \cmidrule(lr){2-9}
        &\textbf{Two-Stage} & & & & & & & \\
        & ChatScene~\cite{chatscene} & 55.5 & 50.2 & 21.6 & 87.7 & 18.0 & 41.6 & 54.6 \\
        & LEO~\cite{leo} & - & - & 24.5 & 101.4 & 20.0 & 49.2 & 50.0 \\
        \textbf{LMMs}
        & Video-3D LLM - 7B~\cite{video3dllm} & 58.1 & 51.7 & 30.1 & 102.1 & - & - & 58.6 \\

        \cmidrule(lr){2-9}
        &\textbf{Single-Stage} & & & & & & & \\
        & NaviLLM~\cite{navillm} & - & - & 23.0 & 75.9 & 15.4 & 38.4 & - \\
        & LL3DA~\cite{ll3da} & - & - & - & 76.8 & 15.9 & 37.3 & - \\
        & SceneLLM~\cite{scenellm} & - & - & 27.2 & 80.0 & 16.6 & 40.0 & 54.2 \\
        & 3D-LLM~\cite{3dllm} & 30.3 & - & 20.5 & 69.4 & 14.5 & 35.7 & - \\
        & Grounded 3D-LLM~\cite{grounded3dllm} & 48.6 & 44.0 & - & 75.9 & - & - & - \\
        & Reason3D~\cite{reason3d} & 49.6 & 41.1 & - & 73.5 & 15.1 & 37.4 & - \\
        & LLaVA-3D - 7B~\cite{l3d} & 50.1 & 42.7 & 27.0 & \textbf{103.1} & \textbf{20.8} & \textbf{49.6} & 55.6 \\
        & \model\;- 3B (Ours) & 62.2 & \textbf{55.3} & 26.6 & 93.4 & 18.3 & 45.7 & 55.8 \\
        & \model\;- 7B (Ours) & \textbf{62.8} & 54.7 & \textbf{27.3} & 96.1 & 19.0 & 46.6 & \textbf{56.8} \\

        \bottomrule
    \end{tabular}}
    \label{table:new_3d_vqa}
    \vspace{-1.0em}
\end{table*}

We evaluate \model{} against existing LMMs and specialized 3D vision models on visual grounding in both in-domain (\cref{sec:3dg}) and out-of-domain settings (\cref{sec:ood_3dg}), along with 3D instance segmentation (\cref{3d_ins}) and 3D VQA (\cref{3dqa}). We also assess how well \model{} retains its 2D multimodal capabilities (\cref{sec:2d_tasks}) and analyze which design choices most significantly impact performance (\cref{sec:analysis}). 
Qualitative results, failure mode analysis, robustness to depth and camera pose noise, and performance on text-only tasks are provided in the appendix. 

\noindent \textbf{Training Datasets} We train jointly on a mixture of 3D and 2D datasets to enable 3D comprehension while preserving the base model's pre-trained capabilities. The 3D datasets include referential grounding (SR3D, NR3D~\cite{referit3d}, ScanRefer~\cite{scanrefer}), instance segmentation (ScanNet200~\cite{scannet200}, Matterport~\cite{matterport}), and question answering (ScanQA~\cite{scanqa}, SQA3D~\cite{sqa3d}). To mitigate catastrophic forgetting of Qwen's original capabilities, we co-train on 2D datasets: referential grounding (RefCOCO, RefCOCO+, RefCOCOg~\cite{refcoco}), instance segmentation (COCO~\cite{coco}), captioning and QA (LLaVA-Instruct-150k~\cite{llava}), and instruction fine-tuning (Alpaca~\cite{taori2023alpaca}).


\label{sec:experiments}
\subsection{Evaluation on 3D Referential Grounding}
\label{sec:3dg}
\textbf{Datasets.} 
We evaluate on the validation sets of three ScanNet-based~\cite{scannet} 3D referential grounding benchmarks: SR3D, NR3D~\cite{referit3d}, and ScanRefer~\cite{scanrefer}. While SR3D comprises 88k synthetic utterances, NR3D (41k) and ScanRefer (51k) feature complex, human-annotated queries. Following recent work~\cite{odin,univlg,locate3d,liftgs}, we operate directly on noisy, raw sensor RGB-D point clouds rather than clean, post-processed meshes. Although this setup introduces sensor-mesh misalignments that can degrade performance~\cite{odin,univlg}, it better reflects practical embodied learning scenarios.


\noindent\textbf{Evaluation Metrics.} 
We report standard Top-1 accuracy, where a prediction is correct if the highest-confidence predicted bounding box achieves an Intersection over Union (IoU) with the ground-truth box above a threshold (0.25, 0.5). 
Since our model predicts segmentation masks, we convert masks to bounding boxes by thresholding at their extreme corners.

\noindent\textbf{Baselines.} 
Following prior work~\cite{l3d,video3dllm}, we compare \model{} against state-of-the-art expert (non-LMM) and LLM-based approaches. Expert baselines include two-stage~\cite{3dvista,pq3d} and single-stage~\cite{locate3d,univlg} methods. LLM-based baselines include: (i) two-stage models relying on detector proposals (LEO~\cite{leo}, Chat-Scene~\cite{chatscene}, Video-3D-LLM~\cite{video3dllm}); (ii) single-stage text-space decoders (3D-LLM~\cite{3dllm}); and (iii) single-stage $<REF>$ token decoders (Grounded 3D-LLM~\cite{grounded3dllm}, LLaVA-3D~\cite{l3d}, Reason3D~\cite{reason3d}). Unlike these, our single-stage method directly decodes segmentation masks by routing LMM backbone features to an object mask decoder. We also evaluate against proprietary 2D VLMs (GPT-4o~\cite{openai2024gpt4o}, LLaMA~\cite{llama}, Qwen2-VL~\cite{qwen2vl}). \cref{table:new_3d_vqa} presents quantitative results, and full results on ReferIt3D are available in the appendix.

\noindent \textbf{\model{} establishes a new state-of-the-art among 3D LMMs.} 
Both our 3B and 7B models surpass the text-decoding single-stage model of 3D-LLM~\cite{3dllm} by over 30\%, the recent single-stage state-of-the-art LLaVA-3D~\cite{l3d} by 12\%, and the two-stage Video-3D-LLM~\cite{video3dllm} by 4\%. This establishes \model{} as the new state-of-the-art for 3D referential grounding among LMM-based models. 




\noindent \textbf{\model{} closes the gap with expert 3D grounding models.}
On ScanRefer~\cite{scanrefer}, \model{} closely matches the state-of-the-art expert model UniVLG, substantially narrowing the gap between specialist 3D models and LMM-based approaches, and outperforming all other methods.


\subsection{Out-of-Domain 3D Referential Grounding}
\label{sec:ood_3dg}
\begin{table}[t]
\centering
\small
\caption{\textbf{Evaluation on Locate3D ScanNet++.}}
\label{tab:scannetpp}
\small
\resizebox{0.70\linewidth}{!}{%
\begin{tabular}{lcc}
\toprule
\textbf{Model} & \textbf{Acc@25} & \textbf{Acc@50} \\
\midrule
UniVLG~\cite{univlg} & 32.3 & 24.6 \\
Video-3D LLM~\cite{video3dllm} & 33.2 & 27.6 \\
\model{}\;- 3B (Ours) & \textbf{55.7} & \textbf{43.4} \\
\bottomrule
\end{tabular}%
}
\end{table}
While LMMs often trail specialists in-domain, they typically excel at out-of-domain (OOD) generalization. We evaluate \model{} on Locate-3D~\cite{locate3d}, which provides human instructions for ScanNet++~\cite{scannetpp} scenes. ScanNet++ introduces a distinct domain shift from our fine-tuning data as it is captured via iPhone LiDAR rather than ScanNet's iPad Structure sensor.


We compare against public checkpoints of UniVLG and Video-3D-LLM (supplied with state-of-the-art ODIN~\cite{odin} box proposals). We omit LLaVA-3D~\cite{l3d} as its grounding model weights and code are not publicly released. As shown in~\cref{tab:scannetpp}, \model{} significantly outperforms both baselines on these OOD tasks. We attribute the improved generalization over UniVLG to \model{}'s stronger pre-training. Furthermore, while Video-3D-LLM uses a large VLM backbone, its reliance on off-the-shelf 3D detectors bottlenecks OOD robustness. Conversely, \model{} directly decodes boxes from VLM features, allowing it to better exploit the underlying representation for superior generalization.

\subsection{3D Instance Segmentation}
\label{3d_ins}
\begin{table}[t]
\centering
\small
\caption{\textbf{Evaluation on ScanNet200 Instance Segmentation.}}
\label{tab:scannet200_ins}
\small
\setlength{\tabcolsep}{3pt}
\resizebox{0.8\linewidth}{!}{%
\begin{tabular}{llcc}
\midrule
 & \textbf{Model}  & \textbf{mAP}  & \textbf{mAP25} \\ \midrule
& Mask3D \cite{mask3d} & 27.4   & 42.3 \\
{Closed}
& PQ3D \cite{pq3d}  & 27.0  & 46.3 \\
{Vocabulary}
& MAFT \cite{maft}  & 29.2  & 43.3 \\ 
& ODIN \cite{odin}  &  \textbf{31.5  }             &  \textbf{53.1} \\  
\midrule 
{\revision{Language-}}
& PQ3D ~\cite{pq3d} & 20.2  & 32.5 \\
{Prompted}
 & UniVLG~\cite{univlg}  &   27.9    & 46.1 \\ 
 \midrule
{\revision{LLM-}}
 & Grounded-3D-LLM~\cite{grounded3dllm} & 12.1 & 16.8 \\
 {Based}
  & \model{}\;- 3B (Ours)  & \textbf{25.3} & \textbf{41.5} \\  \midrule
\end{tabular}%
}
\end{table}

We evaluate \model{} on the ScanNet200~\cite{scannet200} instance segmentation benchmark. While traditional methods~\cite{odin,mask3d} assume a closed vocabulary setup, recent models~\cite{pq3d,univlg}—like ours—adopt a language-prompted paradigm (e.g., ``find chairs. tables. sofa.''). Furthermore, because most 3D grounding VLMs predict only a few bounding boxes, they fail as full scene detectors, with Grounded-3D-LLM~\cite{grounded3dllm} being the only exception to our knowledge. As shown in~\cref{tab:scannet200_ins}, \model{} outperforms Grounded-3D-LLM by 13\% mAP and 25\% mAP25, approaching the performance of language-prompted specialist models.

\subsection{3D Visual Question Answering}
\label{3dqa}
We evaluate \model{} on two 3D question answering benchmarks: ScanQA~\cite{scanqa} and SQA3D~\cite{sqa3d}. Both datasets use visual scenes from ScanNet~\cite{scannet}, with ScanQA focusing on spatial-relation questions and SQA3D emphasizing situational reasoning. 

Following prior work, we report Exact Match (EM@1), ROUGE (R), CIDEr (C), and METEOR (M). As shown in~\cref{table:new_3d_vqa}, \model{} on the 3B variant outperforms state-of-the-art expert model UniVLG~\cite{univlg} on all metrics, and \model{}-7B matches or exceeds the strongest single-stage LMMs on Exact Match. 



\subsection{2D Vision-Language tasks}
\label{2dvl}
\begin{table}[tb]
    \centering
    \caption{\textbf{2D Ref. grounding datasets and RealWorldQA}}
    \vspace{2pt}
    \resizebox{.49\textwidth}{!}
    {
    \begin{tabular}{ll*{3}{c}}
         \toprule
        \textbf{Model} & \textbf{RefCOCO} & \textbf{RefCOCO+} & \textbf{RefCOCOg} & \textbf{RealWorldQA} \\
         \midrule
        LAVT \cite{lavt} (B) & 72.7 & 62.4 & 61.2 & - \\
        ReSTR \cite{rester} & 67.2 & 55.7 & 54.5 & - \\
        X-Decoder (L) \cite{xdecoder} & - & - & 64.6 & - \\
        UniVLG~\cite{univlg} & 69.2 & 61.3 & 64.1 & - \\
        Qwen-2.5-VL-3B
        \cite{qwen2.5}& 89.1 & 82.4 & \textbf{85.2} & \textbf{62.6} \\
        Qwen-2.5-VL-3B (mask decoder) & \textbf{89.7} & \textbf{86.4} & 84.4 & - \\
        \model{}\;- 3B (Ours) & 88.1 & 82.7 & 84.0 & 60.4 \\
         \bottomrule
    \end{tabular}
    }
    \label{table:refcoco}
    \vspace{-0.8em}
\end{table}

\label{sec:2d_tasks}
To prevent degradation of the base model's strong 2D capabilities, we co-train \model{} on 2D datasets, including the RefCOCO family~\cite{refcoco} and LLaVA-Instruct-150k~\cite{llava}. \cref{table:refcoco} evaluates \model{} against its pre-trained base model (Qwen2.5-VL 3B~\cite{qwen2.5}) on RefCOCO 2D grounding benchmarks and the held-out RealWorldQA~\cite{realworldqa} dataset. 
As in our 3D setup, \model{} predicts segmentation masks which we convert to bounding boxes for evaluation. Results show that \model{} preserves the strong 2D capabilities of Qwen2.5-VL despite additional 3D fine-tuning, indicating that geometry-aware reasoning can be added without sacrificing the general multimodal capabilities of the base model.

For additional visualizations of \model{}, ablations on pre-training retention, and results on text-only tasks, please refer to the appendix.

\subsection{Additional Analysis and Ablations}
\label{sec:analysis}
\begin{table*}[ht!]
    \centering
    \caption{\textbf{Ablations of \model{}}}
    \label{table:ablations_qwen3d}
    \scriptsize
    \setlength{\tabcolsep}{2pt}
    \renewcommand{\arraystretch}{1.05}
    \begin{subtable}[t]{0.24\textwidth}
        \centering
        \caption{\textbf{Text vs. $<\mathrm{REF}>$}}
        \label{tab:ref}
        \resizebox{\linewidth}{!}{%
        \begin{tabular}{lc}
            \toprule
            \textbf{Model} & \textbf{Top1 Acc.} \\
            \midrule
            Grounded 3D-LLM & 44.0 \\
            LLaVA-3D - 7B & 42.7 \\
            \model{} $<\mathrm{REF}>$ & 39.7 \\
            \model{} (ours) & \textbf{53.5} \\
            \bottomrule
        \end{tabular}
        }
    \end{subtable}
    \hfill
    \begin{subtable}[t]{0.24\textwidth}
        \centering
        \caption{\textbf{Pos. Embed.}}
        \label{tab:pe}
        \resizebox{\linewidth}{!}{%
        \begin{tabular}{lc}
            \toprule
            \textbf{Pos. Embed.} & \textbf{Top1 Acc}  \\
            \midrule
            2D  & 53.2\\ 
            Vanilla 3D & 49.9 \\
            Aligned 3D (ours) & \textbf{53.5} \\
            \bottomrule
        \end{tabular}
        }
    \end{subtable}
    \hfill
    \begin{subtable}[t]{0.20\textwidth}
        \centering
        \caption{\textbf{Attn. Mask}}
        \label{tab:mask}
        \resizebox{\linewidth}{!}{%
        \begin{tabular}{lc}
            \toprule
            \textbf{Mask Type} & \textbf{Top1 Acc.} \\
            \midrule
            Causal & 36.4 \\
            Full & \textbf{53.5}\\
            \bottomrule
        \end{tabular}
        }
    \end{subtable}
    \hfill
    \begin{subtable}[t]{0.20\textwidth}
        \centering
        \caption{\textbf{VLM Tuning}}
        \label{tab:finetune}
        \resizebox{\linewidth}{!}{%
        \begin{tabular}{lc}
            \toprule
            \textbf{Finetune} & \textbf{Top1 Acc.} \\
            \midrule
            Frozen & 37.0 \\
            Finetune & \textbf{53.5} \\
            \bottomrule
        \end{tabular}
        }
    \end{subtable}
\end{table*}

We ablate our key design choices on ScanRefer (Top-1@0.5) utterances from 50 randomly sampled scenes from the validation set using the 3B variant of our model.
\vspace{-1em}

\paragraph{Geometry-Aware Decoding:} 
We ablate the full token interface between the vision-language attention module and the object mask decoder against a variant that replaces the full text token injection with a special grounding token $<\mathrm{REF}>$ into the mask decoder, similar to LLaVA-3D~\cite{l3d} and Grounded-3D-LLM~\cite{grounded3dllm}. As shown in \cref{tab:ref}, with the $<\mathrm{REF}>$ token, \model{} underperforms both LLaVA-3D and Grounded-3D-LLM, however, \model{} with the full text and visual token interface greatly outperforms all $<\mathrm{REF}>$ token baselines. 

\vspace{-1em}
\paragraph{Geometry-Aware Attention:} We compare the 3D positional embeddings against the 2D MRoPE utilized in Qwen2.5-VL and a naive implementation of 3D RoPE in Qwen's MRoPE on 3D grounding. As shown in~\cref{tab:pe}, our 3D RoPE outperforms all other RoPE variants. We hypothesize this is because 3D RoPE improves the model's cross-view spatial reasoning over the 2D variant, while the distribution of the frequencies in Qwen's original MRoPE hinders the model's capability to discern spatial relationships along the point cloud's axes. We provide further details of the alternative RoPE designs in the appendix. 

\vspace{-1em}
\paragraph{Causal Mask vs. Full Attention Mask:}
Qwen2.5-VL applies causal masking in its vision–language attention layers, even for non-autoregressive tasks such as grounding. We compare this original design—which stays closer to the model’s pre-training distribution—against a full attention mask where all vision and language tokens attend to each other except during autoregressive text generation. As shown in~\cref{tab:mask}, the full-attention variant significantly outperforms the causal-mask variant.
\vspace{-1em}
\paragraph{Freezing vs. Fine-tuning the Base VLM:} 
As shown in \cref{tab:finetune}, fine-tuning the underlying Qwen VLM is essential to achieve strong grounding performance. Keeping the VLM frozen leads to a substantial drop in accuracy. 

\section{Limitations and Future Directions}

While Qwen-3D performs attention in a shared 3D coordinate system, we assume that this coordinate system is largely static over time. Dynamic environments introduce non-rigid motion, object interactions, and topology changes that cannot be captured by a single persistent scene representation. Extending geometry-aware attention and grounding to dynamic 4D scene representations is an important direction for future work.  This will likely require integrating temporal motion representations, dynamic scene reconstruction, or object-centric tracking into the model’s geometric reasoning pipeline. 

While Qwen-3D grounds language directly into 3D scene representations, most tasks considered in this work involve a single grounding step. Many embodied and interactive applications require compositional reasoning over multiple grounded entities, such as spatial comparisons, relational reasoning, and long-horizon instruction following. Integrating geometry-aware grounding with iterative reasoning frameworks may enable richer forms of scene understanding and planning.

Qwen-3D relies on externally estimated depth maps and camera poses to construct its world-space representation. While our experiments demonstrate robustness to substantial depth and pose noise, errors in geometric reconstruction ultimately limit downstream grounding performance. An interesting future direction is to jointly optimize geometry estimation and language-guided scene understanding within a single end-to-end framework.

\section{Conclusion}
\label{sec:conclusion}

We introduced \model{}, a geometry-aware 3D vision–language model that integrates explicit multi-view geometric reasoning into a strong multimodal backbone. By leveraging depth and camera pose to guide token compression and 3D Rotary Positional Embeddings to perform attention in the world  space, \model{} efficiently processes long multi-view video streams while maintaining consistent cross-view spatial understanding.  We further proposed a tightly integrated grounding architecture that directly connects contextualized vision–language features from the backbone to a 3D query-based segmentation decoder, enabling direct language-to-3D alignment.   
Extensive experiments demonstrate that Qwen-3D substantially advances the state of the art among general-purpose 3D LMMs, improving grounding, segmentation, and visual question answering while preserving strong 2D capabilities. More broadly, our results suggest that geometry-aware representations alone are insufficient for high-quality 3D perception. Closing the gap between language reasoning and dense geometric prediction requires equally strong geometry-aware decoding mechanisms. We believe that tightly coupling world-space representations with dense grounding modules provides a promising foundation for future multimodal models that reason seamlessly across images, videos, and 3D environments.


\section{Acknowledgements}
This material is based upon work supported by the National Science Foundation Graduate Research Fellowship Program under Grant No(s) DGE2140739, an NSF Career award, ONR award N00014-23-1-2415, AFOSR Grant FA9550-23-1-0257. Any opinions, findings, and conclusions or recommendations expressed in this material are those of the author(s) and do not necessarily reflect the views of the National Science Foundation. 
Ayush Jain is supported in part by the Meta AI Mentorship Fellowship. The authors thank Gabriel Sarch and Akshan Agrawal for proofreading and helpful discussions.


{
    \small
    \bibliographystyle{ieeenat_fullname}
    \bibliography{main}
}

\clearpage



\section{Appendix}


\subsection{Sensitivity to Pose and Depth Noise}
While \model{} already operates on real-world sensor noise in all experiments reported in the paper, we further stress-test the model under controlled settings with substantial depth and camera pose noise.

Following UniVLG~\cite{univlg}, we simulate depth noise by injecting Gaussian noise with increasing variance into the raw depth maps before unprojection. To simulate additional pose noise, we add Gaussian noise with increasing variance to both the translation and rotation components of the provided camera poses. 

We evaluate on 50 randomly sampled scenes from ScanRefer (Top-1@0.25) against UniVLG, the strongest reported baseline on ScanRefer. We choose UniVLG as the primary baseline because many strong 3D LMMs, such as Video-3D-LLM~\cite{video3dllm}, adopt two-stage pipelines that rely on external object detection models and operate on pre-processed detections at inference time. Fairly evaluating such methods under additional noise would require these external detectors to be re-run on the perturbed point clouds, making comparisons difficult to standardize. Additionally, existing single-stage 3D LMMs either substantially underperform our model or do not release the necessary visual grounding code (e.g., LLaVA-3D~\cite{l3d}).

\begin{figure*}[ht!]
\centering
\begin{subfigure}[t]{0.49\textwidth}
    \centering
    \includegraphics[width=\textwidth]{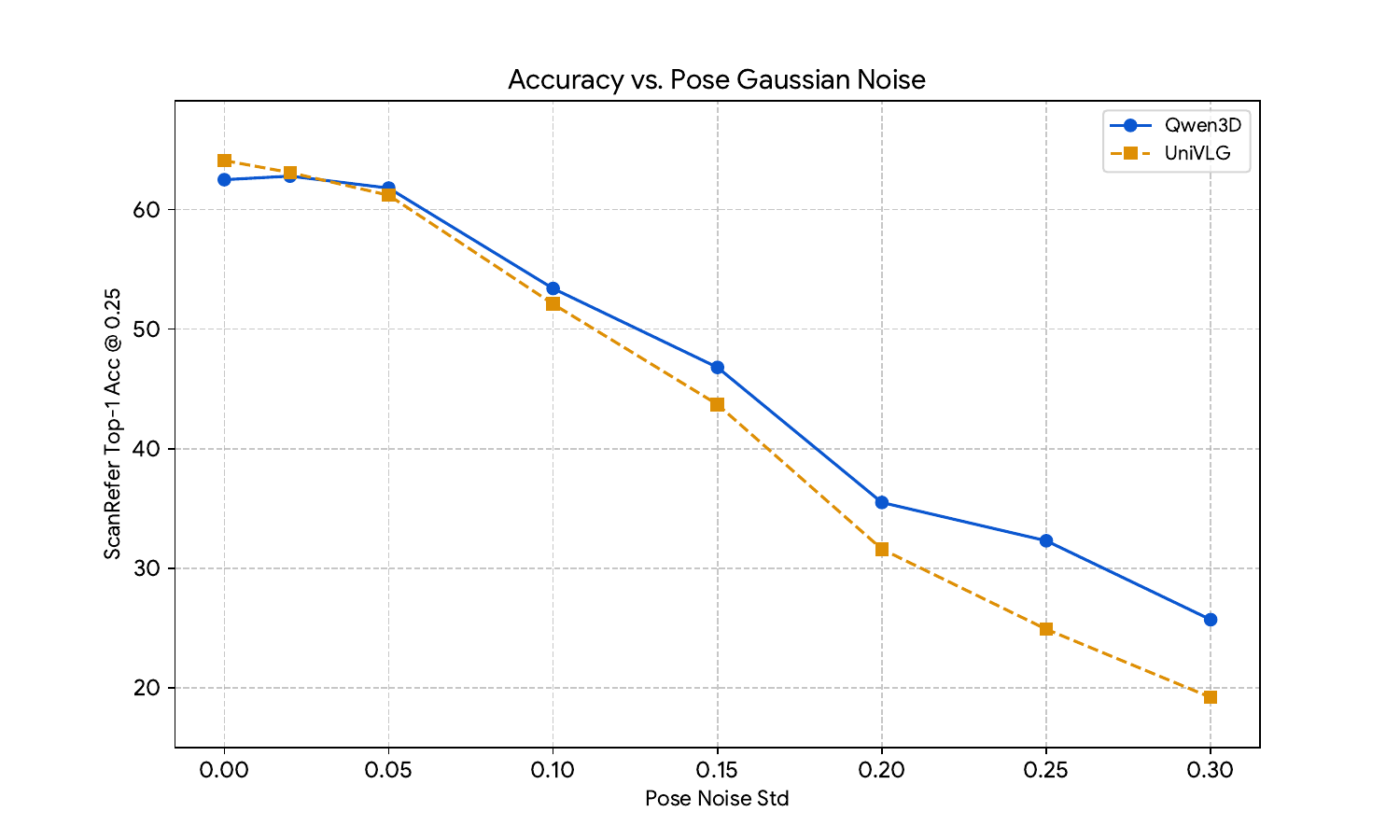}
    \caption{Pose noise sensitivity.}
    \label{fig:pose_noise_sensitivity}
\end{subfigure}\hfill
\begin{subfigure}[t]{0.49\textwidth}
    \centering
    \includegraphics[width=\textwidth]{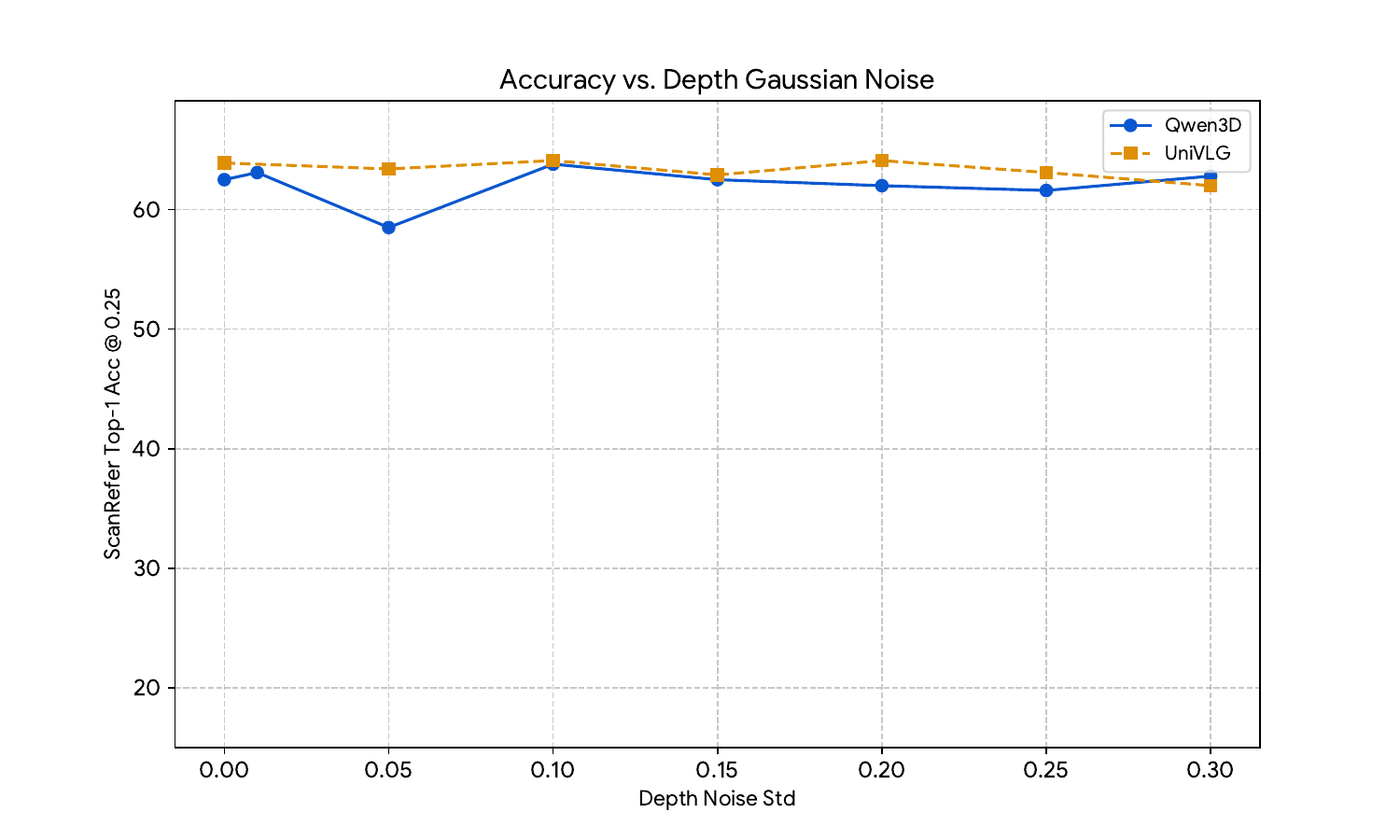}
    \caption{Depth noise sensitivity.}
    \label{fig:depth_noise_sensitivity}
\end{subfigure}
\caption{Sensitivity analysis to camera pose noise (left) and depth noise (right) on ScanRefer (Top-1@0.25). We compare \model{} against state-of-the-art expert model UniVLG~\cite{univlg}}
\label{fig:pose_depth_noise}
\end{figure*}

As shown in~\cref{fig:pose_depth_noise}, \model{} avoids catastrophic failure even under significant noise. Specifically, \model{} demonstrates high robustness to depth noise, showing no significant performance drop even under high variance noise, matching UniVLG's resilience to spurious points. \model{} is similarly robust to pose noise, degrading gracefully and even outperforming the state-of-the-art model under extreme misalignment. We attribute this robustness to the strong integration of 2D pretrained vision-language priors from the Qwen backbone into explicit 3D structures, allowing the model to project reasoning over 2D inputs into accurate 3D masks even when the underlying 3D structures are corrupted.

\subsection{Additional 3D Grounding Results}
We report detailed 3D grounding evaluations on the SR3D, NR3D~\cite{referit3d}, and ScanRefer~\cite{scanrefer} benchmarks in~\cref{table:grounding}. To our knowledge, no prior 3D LLM-based models report on the SR3D and NR3D benchmarks. \model{} outperforms all prior LLM methods on ScanRefer with only a 3B VLM backbone, despite many of the strongest prior methods utilizing more powerful backbones of 7B or more. 

\begin{table*}[!ht]
    \centering
    \caption{\textbf{Results on 3D language grounding for both experts and LMMs.} We evaluate top-1 accuracy on the official validation set. $^*$ UniVLG numbers reproduced using the official code with a single 40GB 8 GPU node instead of their 32 GPU setup, and verified with the authors.}
    \resizebox{.98\textwidth}{!}{
    \begin{tabular}{ll*{11}{c}}
        \toprule
        & & \multicolumn{4}{c}{\textbf{SR3D}} & \multicolumn{4}{c}{\textbf{NR3D}} & \multicolumn{3}{c}{\textbf{ScanRefer}}\\
        & \textbf{Methods} & \begin{tabular}{@{}c@{}}Acc \\ @25\\ (Det)\end{tabular} & \begin{tabular}{@{}c@{}}Acc \\ @50\\ (Det)\end{tabular} & \begin{tabular}{@{}c@{}}Acc \\ @75\\ (Det)\end{tabular} & \begin{tabular}{@{}c@{}}Acc \\ (GT)\end{tabular} & \begin{tabular}{@{}c@{}}Acc \\ @25\\ (Det)\end{tabular} & \begin{tabular}{@{}c@{}}Acc \\ @50\\ (Det)\end{tabular} & \begin{tabular}{@{}c@{}}Acc \\ @75\\ (Det)\end{tabular} & \begin{tabular}{@{}c@{}}Acc \\ (GT)\end{tabular} & \begin{tabular}{@{}c@{}}Acc \\ @25\\ (Det)\end{tabular} & \begin{tabular}{@{}c@{}}Acc \\ @50\\ (Det)\end{tabular} & \begin{tabular}{@{}c@{}}Acc \\ @75\\ (Det)\end{tabular} \\
        \midrule

        & ReferIt3DNet~\cite{referit3d} & 27.7 & - & - & 39.8 & 24.0 & - & - & - & 26.4 & 16.9 & - \\
        & ScanRefer~\cite{scanrefer} & - & - & - & - & - & - & - & - & 35.5 & 22.4 & - \\
        & InstanceRefer~\cite{yuan2021instancerefer} & 31.5 & - & - & 48.0 & 29.9 & - & - & - & 40.2 & 32.9 & - \\
        & LanguageRefer~\cite{roh2022languagerefer} & 39.5 & - & - & 56.0 & 28.6 & - & - & - & - & - & - \\
        & SAT-2D~\cite{yang2021sat} & 35.4 & - & - & 57.9 & 31.7 & - & - & - & 44.5 & 30.1 & - \\
        & BUTD-DETR~\cite{butd} & 52.1 & - & - & 67.0 & 43.3 & - & - & 54.6 & 52.2 & 39.8 & - \\\
        \textbf{Experts}
        & PQ3D~\cite{pq3d} & 62.0 & 55.9 & 46.2 & 79.7 & 52.2 & 45.0 & 37.6 & 66.7 & 56.7 & 51.8 & 43.3 \\
        & 3D-VisTA~\cite{3dvista} & 56.5 & 51.5 & 42.8 & 76.4 & 47.7 & 42.2 & 35.5 & 65.1 & 51.0 & 46.2 & 36.7 \\
        & ODIN~\cite{odin} & 38.1 & 29.3 & 23.1 & - & 31.6 & 20.8 & 15.8 & - & 43.1 & 33.4 & 26.2 \\
        & Locate-3D~\cite{locate3d} & 65.8 & 52.9 & - & - & 53.7 & 40.5 & - & - & 59.9 & 49.6 & - \\
        & Locate-3D+~\cite{locate3d} & 68.2 & 54.8 & - & - & 56.1 & 43.2 & - & - & 61.1 & 50.9 & - \\
        & UniVLG (1 GPU node)~\cite{univlg}$^*$ & 71.7 & 63.6 & 51.2 & - & 52.8 & 42.0 & 33.0 & - & 64.1 & 52.7 & 43.5 \\
        & UniVLG~\cite{univlg} & \textbf{73.0} & \textbf{64.8} & \textbf{51.8} & - & \textbf{58.3} & \textbf{49.8} & \textbf{39.1} & - & \textbf{63.5} & \textbf{56.4} & \textbf{46.0} \\

        \midrule

        & \textbf{2D VLMs}$^*$ & & & & & & & & & & & \\
        & LLaMA~\cite{llama} & 21.3 & 13.9 & - & - & 28.0 & 16.9 & - & - & 37.3 & 24.2 & - \\
        & Qwen2-VL-2B~\cite{qwen2vl} & - & - & - & - & 31.1 & - & - & - & 35.9 & 31.9 & - \\
        & Qwen2-VL-72B~\cite{qwen2vl} & - & - & - & - & 47.6 & - & - & - & 44.8 & 40.3 & - \\
        & Qwen2.5-VL-3B~\cite{qwen2.5llm}  & - & - & - & - & - & - & - & - & 36.4 & 11.8 & - \\
        & Qwen2.5-VL-7B~\cite{qwen2.5llm}  & - & - & - & - & - & - & - & - & 41.6 & 14.9 & - \\
        & GPT-4o~\cite{openai2024gpt4o} & 29.2 & 18.9 & - & - & 38.2 & 25.1 & - & - & 48.2 & 32.5 & - \\

        \cmidrule(lr){2-13}
        & \textbf{Two-Stage} & & & & & & & & & & & \\
        & ChatScene~\cite{chatscene} & - & - & - & - & - & - & - & - & 55.5 & 50.2 & - \\
        & LEO~\cite{leo} & - & - & - & - & - & - & - & - & - & - & - \\
        \textbf{LMMs}
        & Video-3D LLM - 7B~\cite{video3dllm} & - & - & - & - & - & - & - & - & 58.1 & 51.7 & - \\

        \cmidrule(lr){2-13}
        & \textbf{Single-Stage} & & & & & & & & & & & \\
        & NaviLLM~\cite{navillm} & - & - & - & - & - & - & - & - & - & - & - \\
        & LL3DA~\cite{ll3da} & - & - & - & - & - & - & - & - & - & - & - \\
        & SceneLLM~\cite{scenellm} & - & - & - & - & - & - & - & - & - & - & - \\
        & 3D-LLM~\cite{3dllm} & - & - & - & - & - & - & - & - & 30.3 & - & - \\
        & Grounded 3D-LLM~\cite{grounded3dllm} & - & - & - & - & - & - & - & - & 48.6 & 44.0 & - \\
        & Reason3D~\cite{reason3d} & - & - & - & - & - & - & - & - & 49.6 & 41.1 & - \\
        & LLaVA-3D - 7B~\cite{l3d} & - & - & - & - & - & - & - & - & 50.1 & 42.7 & - \\
        & \model\; - 3B (Ours) & \textbf{60.4} & 53.4 & \textbf{40.6} & - & 57.6 & \textbf{49.7} & \textbf{39.1} & - & 62.2 & \textbf{55.3} & \textbf{44.1} \\
        & \model\; - 7B (Ours) & 59.0 & \textbf{53.7} & 37.0 & - & \textbf{58.0} & 49.1 & 38.9 & - & \textbf{62.8} & 54.7 & 43.7 \\

        \bottomrule
    \end{tabular}}
    \label{table:grounding}
\end{table*}

\begin{table}[t]
\centering
\small
\caption{\textbf{Zero-shot evaluation on 3EED.}}
\label{tab:3eed_evaluation}
\small
\setlength{\tabcolsep}{3pt}
\resizebox{0.70\linewidth}{!}{%
\begin{tabular}{llcc}
\midrule
 & \textbf{Model} & \textbf{A@0.25} & \textbf{A@0.50} \\ \midrule
{\textit{Drone}}
& UniVLG~\cite{univlg} & 14.6 & 8.4 \\
& \model{}\;- 3B (Ours) & \textbf{17.8} & \textbf{9.0} \\
\midrule
{\textit{Quad}}
& UniVLG~\cite{univlg} & 10.6 & 3.1 \\
& \model{}\;- 3B (Ours) & \textbf{14.2} & \textbf{5.1} \\
\midrule
{\textit{Vehicle}}
& UniVLG~\cite{univlg} & \textbf{28.6} & \textbf{11.5} \\
& \model{}\;- 3B (Ours) & 22.9 & 10.2 \\
\midrule
\end{tabular}%
}
\end{table}

\subsection{Out-of Domain Outdoor 3D Referential Grounding}
We evaluate our model out-of-domain on the 3EED~\cite{li20253eed} outdoor 3D referential grounding benchmark. Since the benchmark utilizes sparse lidar point clouds not aligned with their provided RGB images, we use MapAnything to perform metric depth completion and obtain dense, metric depth maps for each RGB image. We also use Segment Anything to obtain bounding boxes aligned to the RGB images and project them to 3D using this aligned depth to obtain 3D ground truth bounding boxes aligned with the dense RGB-D point clouds. We evaluate in the same setup against UniVLG and find that our model matches UniVLG overall and outperforms it on the drone and quad splits. We exclude Video-3D-LLM since it relies on object detectors typically trained on indoor data that does not cover the classes used in this benchmark.  

\subsection{Test-Time Frame Ablation}
\begin{figure*}[ht!]
\centering
\includegraphics[width=0.8\textwidth]{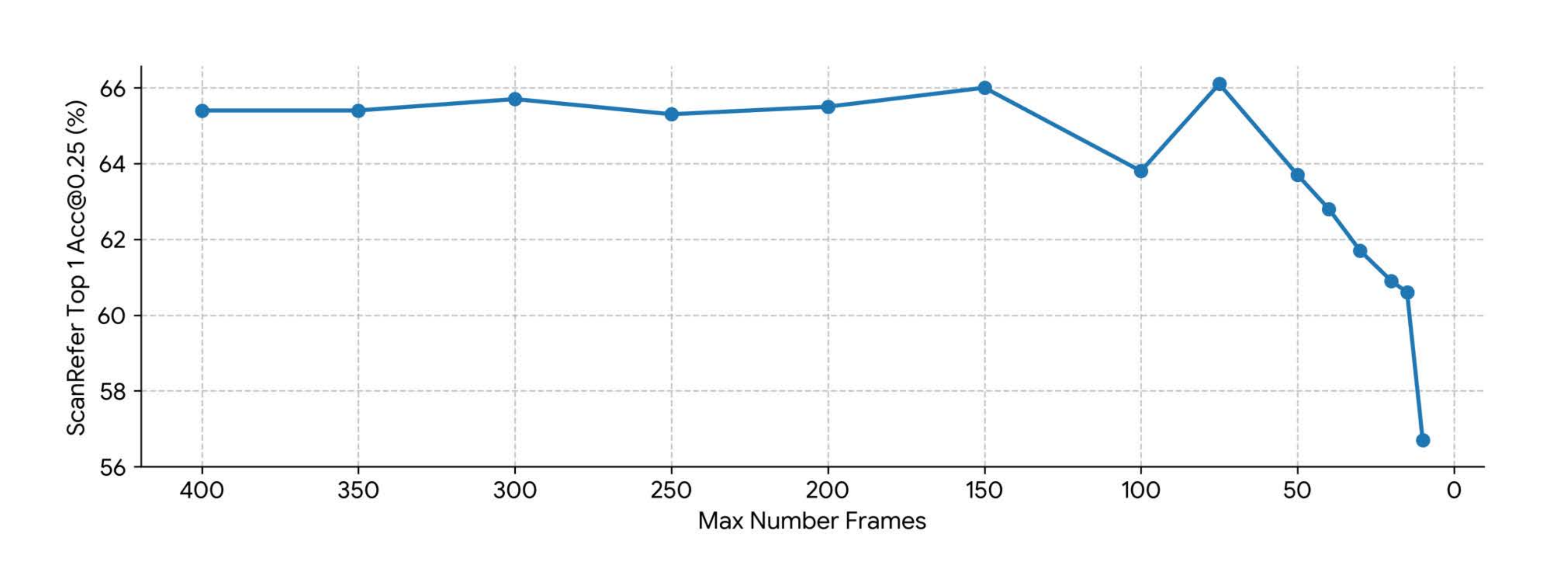}
\caption{Sensitivity to number of views on ScanRefer (Top-1@0.25)}
\label{fig:num_views}
\end{figure*}

We conduct an ablation on the number of input frames at test-time on 50 randomly sampled scenes from ScanRefer, gradually decreasing the maximum number of input frames. We find that even when decreasing the number of views drastically from a maximum of 400 frames to 15 frames, \model{}'s grounding performance does not degrade catastrophically, maintaining a grounding performance of 56\% even with 15 input views in ~\cref{fig:num_views}. 

\subsection{End-to-end Latency}
We estimate the end-to-end latency of \model{} on 3D tasks, tracking from raw scene inputs to the final 3D grounding outputs. As noted in Section 3 of the main text, doing an inference on the full set of multi-view RGB-D frames through \model{} takes approximately 2 seconds. This includes the time required to unproject RGB-D frames (20 ms) and voxelize the 3D point cloud (1 ms). Standard RGB-D SLAM reconstruction methods~\cite{dai2017bundlefusion} used for ScanNet scenes add about 1 to 5 seconds, yielding a total end-to-end latency of roughly 3 to 7 seconds. Notably, SLAM reconstruction costs can be amortized per scene as the reconstruction only needs to be run once; subsequent queries to the same environment only require the 2-second model forwarding time. Additionally, we note that all 3D LMMs need to run SLAM to obtain their point clouds; \model{} does not introduce any new dependencies over prior methods.

\subsection{Detailed Comparison of Training Datasets}
\model{} is trained on a mix of 2D and 3D datasets. The 3D datasets comprises approximately 255k samples spanning 3D question answering~\cite{sqa3d, scanqa}, referential grounding~\cite{referit3d,scanrefer}, and instance segmentation~\cite{scannet200}. The 2D datastes include referential grounding (RefCOCO, RefCOCO+, RefCOCOg~\cite{refcoco}), instance segmentation (COCO~\cite{coco}), captioning and question answering (LLaVA-Instruct-150k~\cite{llava}), and instruction fine-tuning (Alpaca~\cite{taori2023alpaca}). Notably, all of these 2D datasets are included in the pre-training dataset of many modern 2D VLMs, including Qwen2.5VL, the backbone we use. We include them to avoid catastrophic forgetting of 2D capabilities as our model's VLM backbone as it trains on new 3D data. 

Our most competent expert model baseline, UniVLG~\cite{univlg}, is trained on a nearly identical dataset mixture, excluding only LLaVA-Instruct-150k and the text-only Alpaca dataset. In contrast, the state-of-the-art single-stage 3D LMM, LLaVA-3D~\cite{l3d} utilizes a much larger 3D finetuning dataset, compiling over 860K 3D visual-reasoning samples together from various benchmarks alongside the 2D LLaVA-Instruct-150k dataset. The leading two-stage 3D LMM, Video-3D-LLM~\cite{video3dllm}, uses a 3D training mixture similar to ours - substituting Referit3D~\cite{referit3d} with Multi3DRefer~\cite{multi3drefer}. However, it also relies heavily on external object detectors, which are typically trained on ScanNet200 instance segmentation datasets.


\subsection{Rotary Position Embedding Implementation}
To process interleaved visual and textual data in the vision-attention layers, Qwen2.5-VL~\cite{qwen2.5} utilizes a Multimodal Rotary Positional Embedding (mRoPE). This method decomposes positional encodings into temporal, height, and width components. Each token $i$ in a sequence of length $K$ is assigned a position ID in each dimension, and the final positional embedding is the concatenation of the sinusoidal encodings of these components:

\[
\text{PE}(\mathbf{p}) = \text{PE}(t_i; h_i; w_i)_{i=1}^{K},
\]

For text tokens, the temporal ID $t$ increments linearly, while $h$ and $w$ are assigned the same values as $t$, effectively reducing mRoPE to 1D-RoPE. For image tokens, $t$ is held constant while $(h, w)$ are assigned to corresponding pixel coordinates. To maintain temporal consistency across all modalities and dimensions, the position IDs of each modality are incremented by the maximum position ID of the preceding modality, ensuring visual features and text instructions occupy distinct, non-overlapping positions in the positional embeddings.

\noindent\textbf{2D RoPE Adaptation:} This extension aims to stay as close as possible to the Qwen2.5-VL pre-training distribution. Temporal IDs follow the original scheme, while the height and width IDs for voxelized point-cloud tokens are obtained by sampling corresponding 2D pixel coordinates from one of the views where the voxel is visible.




\noindent\textbf{Naive 3D MRoPE:} 
This extension aims to inject absolute 3D spatial information by replacing the height and width components with embeddings of absolute XYZ coordinates: 

\[
\text{PE}(\mathbf{p}) = \text{PE}(t_i; x_i; y_i, z_i)_{i=1}^{K},
\]

This method is similar to our final 3D RoPE, except it utilizes Qwen2.5-VL's partitioning of the feature embedding channels among the four dimensions, resulting in an imbalanced frequency spectrum across the four dimensions, with high-frequency channels being assigned to the temporal axis and low-frequency channels being assigned to the z-axis ~\cite{multimodalpe}. We find that this uneven allocation degrades the model's 3D grounding capabilities in our ablations. 

Our 3D RoPE instead initializes the same range of frequencies for each dimension, resulting in a uniform distribution of high and low frequency channels across all components. We find that this method significantly outperforms all other variants on 3D grounding tasks. 




\subsection{Additional Hyperparameters}
Hungarian Matching cost weights during training have a sizable impact on our model's performance on 3D grounding tasks. Our final configuration utilizes the standard cost coefficients from Mask2Former~\cite{m2f}, as we found that alternative weighting schemes proposed in other 3D grounding works ~\cite{univlg} can hinder performance. 

\subsection{Comparison with Token Merging} 
While \model{} utilizes voxelization pooling to reduce multi-view redundancy, a popular alternative for reducing redundancy in video model architectures is token merging~\cite{bolya2022tome}, which iteratively fuses visually similar tokens within ViT attention layers. We attempted to compare these methods to our voxelization pooling by replacing the voxelization with a plug-and-play token merging strategy within our vision-language attention layers~\cite{bolya2022tome}. This implementation proved infeasible for dense 3D scenes, as with token merging, the first vision-language attention layer must still process the point cloud at full resolution (averaging around 33K points for ScanNet scenes). This consistently resulted in GPU out-of-memory errors during inference, regardless of how aggressively the token merging would downsample in subsequent layers. Conversely, our voxelization pooling preemptively downsamples the point cloud, reducing the number of points to an average of about 14K points prior to the attention layers. This acts as a natural, 3D-aware compression mechanism that bypasses the memory overhead of full-resolution multi-view attention.




\subsection{Additional Implementation Details}

For our model, we use Qwen2.5-VL as the backbone due to its open-source implementation and its tight integration of visual and textual features in its visual-language attention layers. Within our codebase, we also introduce several engineering modifications to the original Qwen2.5-VL architecture.


The Qwen2.5-VL vision encoder handles the batching of multiple images by sequentially concatenating images along the token dimension and using attention masks to prevent cross-image interaction, resulting in $O(N^2)$ complexity in the vision encoder with respect to the number of images $N$. We optimize this by implementing batched forwarding in the attention mechanism to perform per-image batched attention. This modification reduces the complexity from $O(N^2)$ to $O(N)$ while preserving the original behavior of the vision encoder. This also substantially reduces GPU memory usage, which is essential for processing scene videos with large numbers of input RGB images.

Additionally, Qwen2.5-VL utilizes causal masking in its visual-language attention layers, allowing each vision and text token to attend only to itself and to earlier tokens in the sequence. This is problematic for our point cloud inputs because this operation is not permutation-invariant and would cause the model to learn artifacts of the arbitrary ordering of the point cloud sequences rather than the underlying geometry. We rectify this by replacing this causal-masked attention with all-to-all attention for all text and visual inputs and finetuning our backbone with causal masking only over the answer tokens. We ablate these changes in the main text. 



\subsection{Visualizations of Failure cases}
\begin{figure*}[ht!]
\centering
    \includegraphics[width=\textwidth]{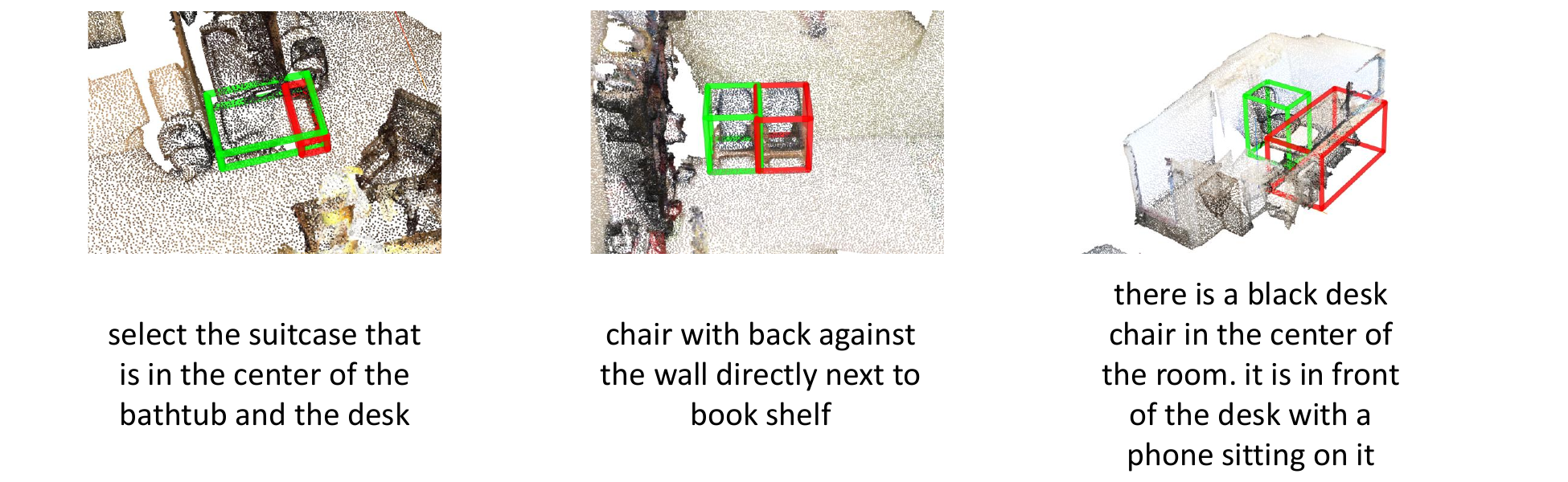}
    \caption{Failure cases of \model{} on 3D grounding tasks. The red segmentation masks and boxes refer to \model{}'s prediction and the green masks and boxes indicate the ground truth.}
    \label{fig:failure}
\end{figure*}

In~\cref{fig:failure}, we outline three common failure modes for \model{} on 3D referential grounding tasks:

\noindent\textbf{Incomplete object selection (Left):} The model occasionally masks only a partial section of the target. This typically occurs when the underlying point cloud contains significant holes or artifacts, skewing the geometric understanding of the complete object shape.

\noindent\textbf{Confusion between instances (Middle):} The model correctly identifies the object class but grounds the wrong instance.  Confusion between multiple instances is a common failure mode seen in mask-decoding architectures, and similar issues have been noted in previous works, such as Mask3D~\cite{mask3d} and UniVLG~\cite{univlg}. 

 \noindent\textbf{Language ambiguity (Right):} The model occasionally confuses the target object with other reference objects mentioned in more ambiguous querying expressions.

\subsection{Visualizations of Qwen-3D Results}
\begin{figure*}[ht!]
\centering
    \includegraphics[width=\textwidth]{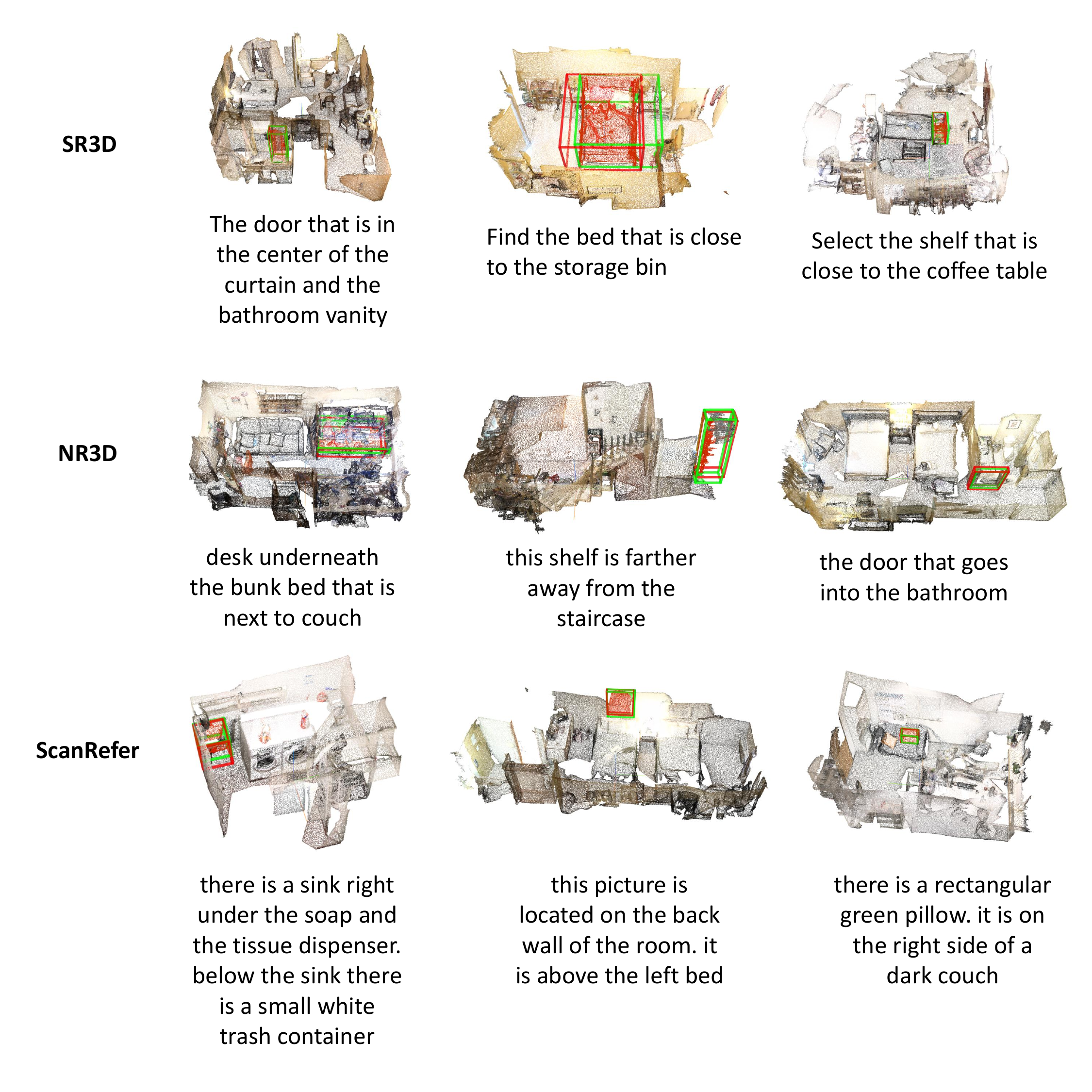}
    \caption{Visualizations of \model{}'s predictions on \textbf{3D Referential Grounding Datasets SR3D, NR3D, and Scanrefer.} The red segmentation masks and boxes refer to \model{}'s prediction and the green masks and boxes indicate the ground truth.}
    \label{fig:3d_ground_viz}
\end{figure*}

\begin{figure*}[ht!]
\centering
    \includegraphics[width=\textwidth]{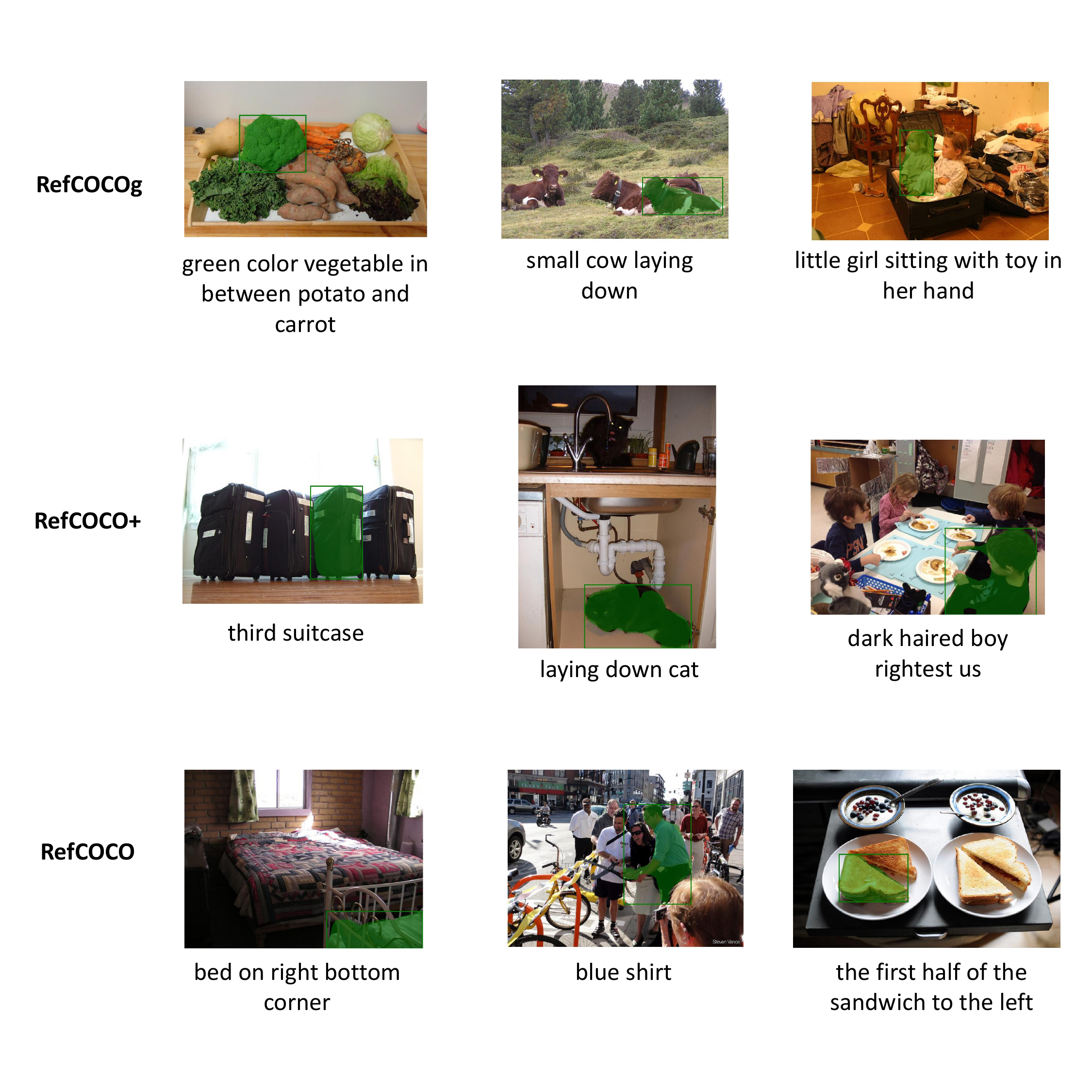}
    \caption{Visualizations of \model{}'s predictions on \textbf{2D Referential Grounding Datasets RefCOCOg, RefCOCO+, and RefCOCO.} The model's predictions are indicated by the green mask and bounding box.}
    \label{fig:2d_ground_viz}
\end{figure*}

\begin{figure*}[ht!]
\centering
    \includegraphics[width=\textwidth]{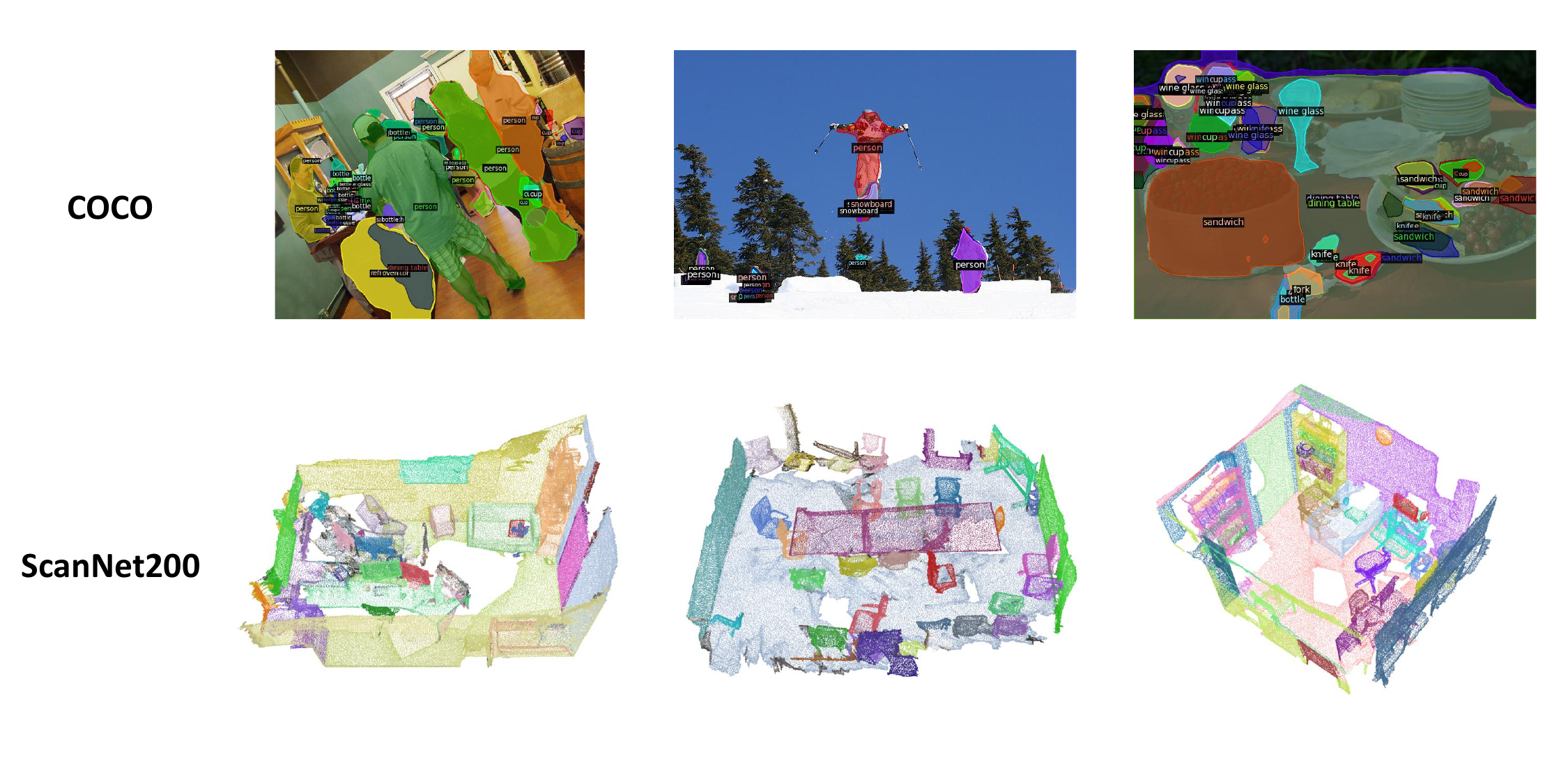}
    \caption{Visualizations of \model{}'s instance segmentation predictions on \textbf{COCO and ScanNet200}.}
    \label{fig:detection}
\end{figure*}

\begin{figure*}[ht!]
\centering
    \includegraphics[width=\textwidth]{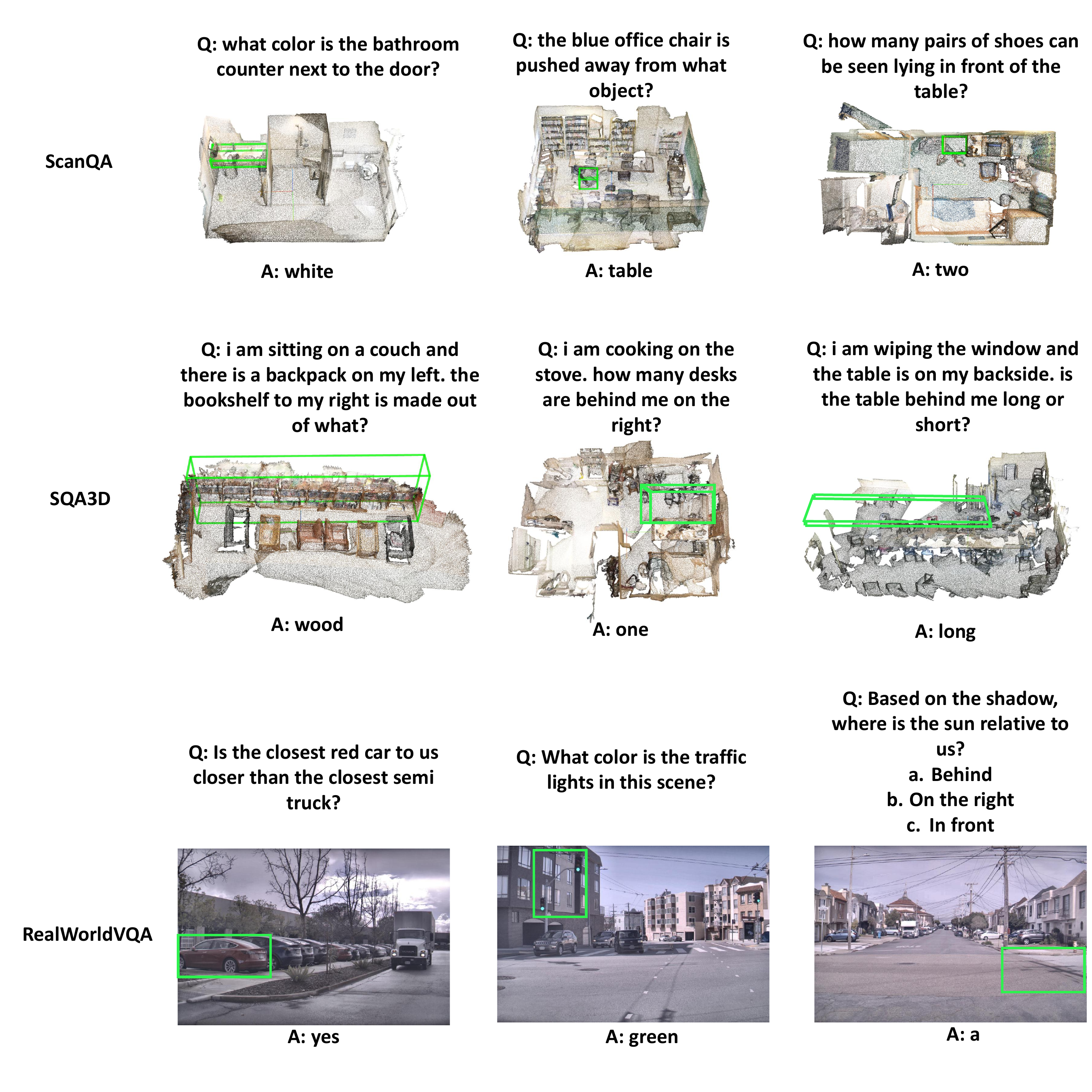}
    \caption{Visualizations of \model{}’s responses to visual question-answering tasks on \textbf{3D benchmarks SQA3D and ScanRefer and the 2D benchmark RealWorldVQA}. Green boxes denote objects relevant to the posed question.}
    \label{fig:vqa}
\end{figure*}

We include visualizations of \model{} predictions in 3D referential grounding tasks in ~\cref{fig:3d_ground_viz}, 2D referential grounding in ~\cref{fig:2d_ground_viz}, instance segmentation in ~\cref{fig:detection}, and visual question-answering tasks in ~\cref{fig:vqa}.

\end{document}